\documentclass[10pt,twocolumn,letterpaper]{article}

\usepackage{iccv}              

\usepackage{xcolor}

\makeatletter
\newcommand\subsubsubsection{\@startsection{paragraph}{4}{\z@}%
            {-2.5ex\@plus -1ex \@minus -.25ex}%
            {0.5ex \@plus 0ex}%
            {\normalfont\normalsize\bfseries}}
\renewcommand{\paragraph}{%
  \@startsection{subparagraph}{5}%
  {\z@}{0.25ex \@plus 1ex \@minus .2ex}{-1em}%
  {\normalfont\normalsize\bfseries}%
}

\makeatother
\definecolor{iccvblue}{rgb}{0.21,0.49,0.74}
\usepackage[pagebackref,breaklinks,colorlinks,allcolors=iccvblue]{hyperref}

\def\paperID{13671} 
\def\confName{ICCV}
\def\confYear{2025}

\title{SAFT: Shape and Appearance of Fabrics from Template \\ via Differentiable Physical Simulations from Monocular Video}

\author{David Stotko
\hspace{25mm}
Reinhard Klein
\\[4mm]
University of Bonn
}

\begin{document}
\maketitle

\begin{abstract}
The reconstruction of three-dimensional dynamic scenes is a well-established yet challenging task within the domain of computer vision.
In this paper, we propose a novel approach that combines the domains of 3D geometry reconstruction and appearance estimation for physically based rendering and present a system that is able to perform both tasks for fabrics, utilizing only a single monocular RGB video sequence as input.
In order to obtain realistic and high-quality deformations and renderings, a physical simulation of the cloth geometry and differentiable rendering are employed.
In this paper, we introduce two novel regularization terms for the 3D reconstruction task that improve the plausibility of the reconstruction by addressing the depth ambiguity problem in monocular video.
In comparison with the most recent methods in the field, we have reduced the error in the 3D reconstruction by a factor of $2.64$ while requiring a medium runtime of $30\,\mathrm{min}$ per scene.
Furthermore, the optimized motion achieves sufficient quality to perform an appearance estimation of the deforming object, recovering sharp details from this single monocular RGB video.
\end{abstract}    
\section{Introduction}

Shape-from-template (SfT) methods offer an elegant solution for reconstructing the 3D geometry of a scene by leveraging prior information in the form of a template and minimal setup requirements, such that a single RGB video is sufficient to tackle reconstruction tasks.
We focus on the challenging task of reconstructing fabrics that deform due to external influences such as wind or human manipulation.
While recent work has shown that incorporating physical simulations enhance the reconstruction process by enforcing realistic motion~\cite{Phi-SfT,stotko2024physics}, such simulations fall short in addressing depth ambiguity where multiple plausible deformations can result in identical RGB renderings.
To address this challenge, we introduce two novel regularization terms that mitigate depth ambiguity, significantly improving the plausibility and quality of the reconstructed geometry.
Our approach is evaluated against state-of-the-art methods, demonstrating superior reconstruction quality with a moderate runtime.

Furthermore, we demonstrate that, in contrast to previous methods, our precisely reconstructed geometry allows the estimation of a highly detailed Spatially Varying Bidirectional Reflectance Distribution Function (SVBRDF) together with the ambient lighting to reconstruct the appearance of the fabric.
Correctly reconstructed dynamic deformations create slightly varying viewing angles and help to decouple diffuse, specular, and lighting contributions.
In summary, we introduce a method that seamlessly combines 3D reconstruction of dynamic non-rigid objects with appearance and illumination reconstruction into a single pipeline, leveraging only a single monocular RGB video.
We will present our results on a real dataset where no ground truth material properties are known, evaluate the quality using synthetic scenes and show the improvements compared to the blurred estimates when using the erroneous geometry of previous techniques~\cite{Phi-SfT,stotko2024physics}.

\urlstyle{same}
The source code is available at \url{https://github.com/vc-bonn/saft}.

\section{Related Work}
\label{section:related_work}

\begin{figure*}
   \centering
   \includegraphics[width=0.99\linewidth]{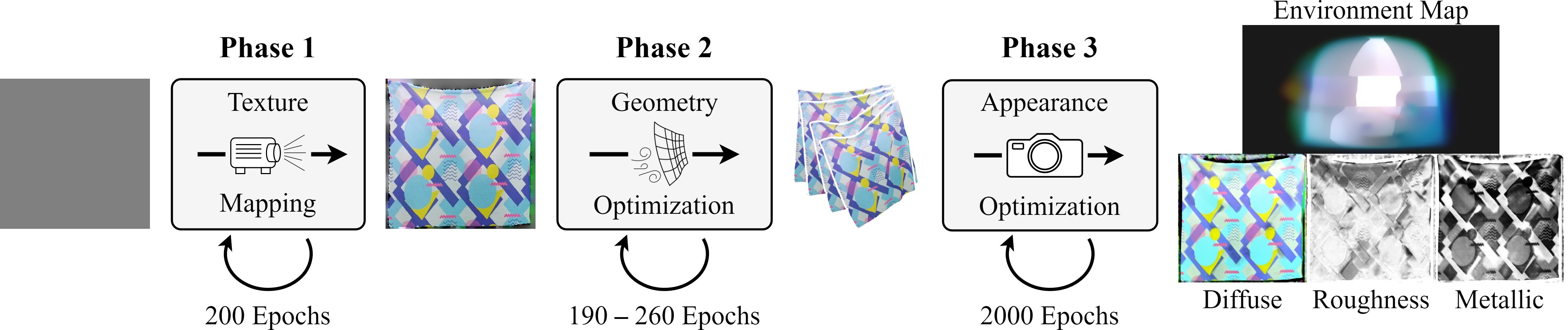}
   \caption{Overview of the three phases of our method. First, we perform projective texture mapping to obtain an accurate texture on the geometry template. Secondly, the precise geometry is reconstructed by optimizing the parameters of a physical simulation to fit rendered images to the target video sequence. Finally, the 3D deformation achieves sufficient quality that it can be used to reliably estimate fine details in appearance parameters and environment lighting from only one monocular RGB video by employing differentiable ray tracing.}
   \label{figure:pipeline_phases}
\end{figure*}

\paragraph{Non-rigid object reconstruction}

The presence of deformable objects introduces various challenges in the context of 3D object reconstruction tasks.
A variety of algorithms are proposed based on the available data and the assumptions made.
Non-rigid structure from motion (NRSfM) algorithms employ one or more cameras to reconstruct arbitrary deformed surfaces~\cite{DDD,Neural-NRSfM} or specialize in human faces~\cite{non_rigid_sfm,Parashar_2020_CVPR} and thin objects like fabric~\cite{HDM-Net,IsMo-GAN}.
Higher quality can be achieved when using some template geometry, potentially including masks, material, textures, or other information to constrain the problem~\cite{deformable_surfaces_1}.
Such methods are called Shape-from-Template (SfT)~\cite{SfT_original,ROBUSfT} and incorporate regularizing constraints, \eg restrictions to isometric deformations resulting in a set of partial differential equations~\cite{isowarp,SfT_original}.
More recent approaches model the deformation field by a neural network~\cite{das2024neural,NN-deformation,NN-SfT} and therefore get smooth and consistent motion.
For a more detailed overview on non-rigid scene reconstruction, we refer to~\cite{reconstruction_overview,reconstruction_overview_old}.

\paragraph{Appearance estimation}

The task of optical material reconstruction heavily depends on the amount of information available.
We want to focus on the estimation of SVBRDFs or parts of it for appearance reconstruction.
Learning-based approaches are able to estimate the SVBRDF of a planar surface from a single image under various lighting conditions \cite{SVBRDF_1,SVBRDF_2,SVBRDF_3,SVBRDF_4,SVBRDF_5,SVBRDF_6}.
Other methods additionally estimate a static object geometry at the same time but require very specific lighting conditions to work \cite{SVBRDF_7,SVBRDF_8,SVBRDF_9}.
High-quality reconstructions can be achieved via inverse rendering techniques if multi-view data is available \cite{Kaltheuner2023Unified,Kaltheuner2025ROSA,iron,nvdiffrec,nvdiffrecmc}.
Some methods also consider to leverage multiple frames of dynamic scenes from a single camera to optimize the diffuse texture only \cite{SVBRDF_dynamic_1,SVBRDF_dynamic_2} or the full SVBRDF~\cite{SVBRDF_dynamic_3}.
Note that all these approaches need RGBD data and are therefore able to work with ground truth geometry.
A more detailed overview of SVBRDF acquisition can be found in the recent survey~\cite{SVBRDF_survey}.

\paragraph{Physics simulation}

Physics simulations are an important part of various different tasks to get realistic dynamics with high precision.
There are several frameworks that support general differentiable computations \cite{DiffTaichi,DiffPD,nvidia_warp} for a variety of systems such as rigid bodies, soft bodies, and fluids.
The simulation of cloth can be realized by using a mass-spring model \cite{baraff_witkin_large_steps,Cirio_yarn_1,Cirio_yarn_2,Gong} or continuum mechanics~\cite{arcsim,Liang_1,Liang_2,DiffCloth}.
The idea of using a differentiable physics simulation \eg as a regularization is already established and yields successful results in various tasks like the reconstruction of humans, animals, and objects 
\cite{Tripathi_2023_CVPR,Yuan_2023_ICCV,Xu_2023_ICCV,PPR,3D_modeling} or estimating cloth parameters~\cite{DiffCloud,Gong,Phi-SfT}.
The major downside of these simulations is their computational complexity which often dominates the overall runtime.

Neural simulation techniques are able to yield high-quality simulations very quickly \cite{PINN,neuralODE}.
Similar to classical simulations, there are general \cite{wandel2024metamizer,NN_mesh_simulation} as well as specialized simulations for fluids \cite{geneva_fluid,wandel_2021}, soft bodies \cite{NN_mesh_simulation}, and cloth in particular \cite{PBNS,SNUG,multi_layer_cloth_NN,NeuralClothSim,NeuralClothSim2}.
Neural networks with physics-based regularizations are used in reconstruction tasks of loose cloth \cite{stotko2024physics}, clothes on human bodies \cite{HOOD,drapenet,caphy}, and parameter estimation \cite{bending_estimation,Yang_2017_ICCV,static_drape_estimation,how_will_it_drape_like} as well.

\paragraph{Similar methods}

Our approach of reconstructing the geometry first (ensuring physical plausibility) and estimating appearance parameters afterwards shares notable similarities with Gaussian Garments~\cite{Gaussian_garments} and PhysAvatar~\cite{PhysAvatar}, despite their reliance on multi-view data.
Better Together~\cite{faceSFS} incorporates shading information directly into the geometry deformations of faces.
The work of Tan~\etal~\cite{SfT_silhouette} tackles the SfT problem by restricting the template to align with the contours in image space and smoothing the geometry using a physics-based deformation regularization.
$\phi$-SfT~\cite{Phi-SfT} uses a sophisticated physics simulation to reconstruct the deformation of cloth.
In contrast, a neural cloth simulation was introduced to reduce the runtime by a factor of $400$ at the expense of slightly worse reconstruction quality and the lack of high-frequency details~\cite{stotko2024physics}.

\section{Method}

\begin{figure*}
   \centering
   \includegraphics[width=0.9\linewidth]{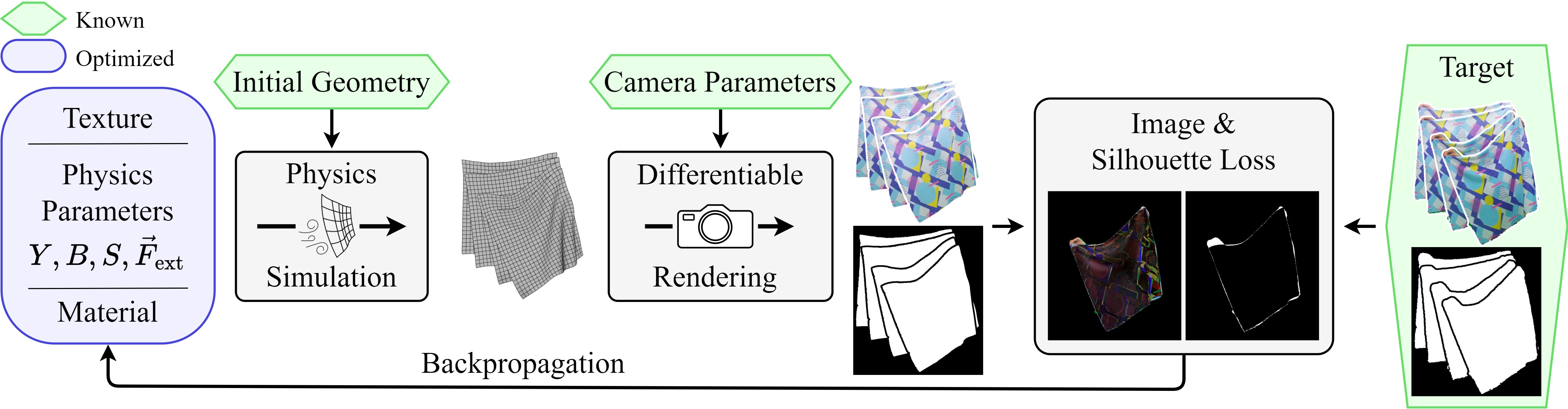}
   \caption{Overview of the pipeline during one epoch. Initially, we perform a physical simulation to obtain the geometry for all frames currently in consideration. Then, renderings and mask images of the fabric are created. Depending on the phase, we either use a texture file for a differentiable rasterizer or an SVBRDF model for a differentiable ray tracer. The images are then compared to the target sequence to define a loss function and perform a gradient-based optimization scheme.}
   \label{figure:pipeline_epoch}
\end{figure*}

Our goal is to perform a 3D reconstruction that produces highly accurate geometry, which facilitates subsequent tasks such as appearance estimation and ensuring the faithful capture of even fine visual details.
We assume our target video sequence to consist of RGB images and corresponding masks, which may be obtained by systems like Segment Anything~\cite{SAM1,SAM2}.
Moreover, we require the initial geometry as a template mesh to start our reconstruction pipeline.
Our approach can be split into three different phases for texture mapping, geometry reconstruction, and appearance estimation as shown in \Cref{figure:pipeline_phases}.
Within these phases, we make use of a physical cloth simulation and differentiable rendering techniques to define the pipeline depicted in \Cref{figure:pipeline_epoch}.
First, \Cref{section:physics,section:rendering} describe the simulation and rendering techniques in more detail.
Then, \Cref{section:phases} presents the phases including their requirements and their outcome.
More details on the simulation and the optimization can be found in the supplementary material.

\subsection{Physical Cloth Model}
\label{section:physics}

We employ a physics simulation as our underlying deformation model because it naturally restricts any dynamics to become physically plausible.
In particular, the cloth is modeled as a mass-spring system, which is on one hand easy to implement and on the other hand still capable of generating highly realistic motion~\cite{Cirio_yarn_1,Cirio_yarn_2,Gong}.

\subsubsection{Equations of Motion}

We represent the cloth as a quad-mesh containing several vertices and edges that also resemble the connections for the mass-spring system.
Hence, each vertex is considered to be a point-mass $M = m I$ with position $\vec{x}(t)$, velocity $\vec{v}(t)$, and acceleration $\vec{a}(t)$.
The relation between these quantities and the superposition of all forces $\vec{F}$ acting on a vertex is given by Newton's second law of motion:
\begin{equation}
\label{equation_of_motion}
    M \vec{a}(t) = M \frac{\partial^2 \vec{x}(t)}{\partial t^2} = \vec{F}(t, \vec{x}, \vec{v})
\end{equation}

\subsubsection{Numerical Integration Scheme}

We employ the backward Euler integration scheme \cite{baraff_witkin_large_steps} to solve \Cref{equation_of_motion} numerically due to its stability with relatively large time steps compared to other integration schemes.
By separating the equations of motion into two first-order differential equations and applying the backward Euler method, we get the update rules
\begin{align}
    \vec{v}_{n+1} & = \vec{v}_n + \Delta t \vec{a}_{n+1} = \vec{v}_n + \Delta t M^{-1} \vec{F}_{n+1} \\
    \vec{x}_{n+1} & = \vec{x}_n + \Delta t \vec{v}_{n+1}
\end{align}
All forces are extrapolated from the current time step by using a first order Taylor approximation to obtain $\vec{F}_{n+1}$.

\subsubsection{Forces and Respective Energies}
\label{section:energies}

The movement of the mesh vertices is entirely described by their initial positions, velocities, and the forces acting on them.
We separate the total forces into several parts, most of which we describe with analytical models.
In particular, we model internal forces for the stretching, bending and shearing of the cloth with simple analytic expressions but also allow for arbitrary external forces $\vec{F}_\mathrm{ext}$ \cite{Phi-SfT,stotko2024physics}.
The internal forces can be modeled using energy terms $E(\vec{x})$ from which the forces acting on a vertex $i$ at position $\vec{x}^i$ are calculated via $\vec{F} = -\frac{\partial E}{\partial \vec{x}^i}$.
We rely on a simple mass-spring model for the energy terms similar to other simulations~\cite{Cirio_yarn_1,Cirio_yarn_2,Gong}. 

\paragraph{Energy Terms}
We defined the stretching energy as the Hookean energy term
\begin{equation}
    E_Y = \frac{1}{2} Y \left( \Vert \vec{e}^{\,ij} \Vert - L_0^{ij} \right)^2
\end{equation}
that penalizes deviations of an edge length $\Vert \vec{e}^{\,ij} \Vert$ from its rest length $L_0^{ij}$ scaled by the stretching stiffness $Y$.
The bending and shearing energies model resistance against angular changes between pairs of edges that enclose angles $\theta^{ijk}$ or $\varphi^{ijl}$.
Both energies are defined using the same analytical form
\begin{equation}
    E_B = \frac{1}{2} B \left( \theta^{ijk} - \theta_0^{ijk} \right)^2 \text{ and } E_S = \frac{1}{2} S \left( \varphi^{ijl} - \varphi_0^{ijl} \right)^2
\end{equation}
with their own bending $B$ or shearing $S$ stiffness parameters, respectively.
The stiffness parameters are shared for all stretching, bending, and shearing segments within the cloth to obtain a homogeneous simulation model.

\subsection{Differentiable Rendering}
\label{section:rendering}

\subsubsection{Rasterization}

During the texture mapping and the deformation optimization, we obtain rendered images via the fast differentiable rasterizer nvdiffrast~\cite{nvdiffrast}.
We render a RGB image $I_n^\mathrm{rast}$ and a mask $M_n$ per frame $n$ of the target video sequence using known camera intrinsic and extrinsic parameters.
These images are used to indicate pixel-wise correspondences to guide the necessary motion and therefore do not need to be rendered perfectly.
In fact, the texture mapping phase yields a texture with baked-in shadows and lighting such that we do not need to compute these contributions in the rendering.
The renderings are used to define image-based loss functions for the color information and the silhouette.

\subsubsection{Ray Tracing}

In contrast to the deformation optimization, we switch to the high-quality differentiable ray tracing of nvdiffrecmc~\cite{nvdiffrecmc} to optimize the SVBRDF.
This enables a decomposition into diffuse, roughness, and metallic textures as well as an estimation of the environment lighting.
We intentionally omit the reconstruction of a normal map because of difficulties and ambiguities when optimizing it from only a single viewpoint and relying on approximate geometry.
The reconstructed motion yields slightly different viewing angles of the fabric which enables the distinction between diffuse texture and specular highlights.
However, this highly depends on the deformation relative to the viewing angle and the lighting conditions in the scene.

We rely on existing work that already dealt with the problem of separating the diffuse, roughness and metallic correctly into respective textures ~\cite{Kaltheuner2025ROSA}.
As we do not impose any illumination prior (\eg illumination by a flashlight), we have to employ a general method that suits our use case.
Therefore, we choose to generate the textures by a trainable decoder~\cite{Kaltheuner2025ROSA,Kaltheuner2023Unified}.
It is part of an auto-encoder that is usually pre-trained on a large optical material database but we omitted the pre-training as it did not improve the results in our case.
The decoder does not take any input and provides several texture patches that are blended together to achieve the desired texture resolution.
We choose to generate $16$ patches with $128 \times 128$ resolution and an overlap of $10$ pixels yielding a resolution of $472 \times 472$ for the final output textures. 
We also estimate the illumination in the form of an environment map texture with a resolution of $512 \times 256$ pixels, which is independent of the decoder for the textures.

\subsection{Optimization Phases}
\label{section:phases}

\subsubsection{Texture Mapping}
\label{section:phase_texture_mapping}

We assume to start with a template that represents a ground truth geometry for the first frame of our video sequence.
Hence, the silhouette as well as the texture of a rendered cloth are supposed to align almost perfectly with the target.
In the case of a template without (satisfactory) texture we can utilize this information to perform a projective texture mapping.
In particular, we optimize the cloth texture $T$ by minimizing the combined loss
\begin{equation}
    \mathcal{L}_\mathrm{tex} = \mathcal{L}_{0,\mathrm{im}} + \lambda_T \mathcal{R}_T
\end{equation}
with $\mathcal{L}_\mathrm{0,im} = \frac{1}{N_\mathrm{p}} \Vert I_0^\mathrm{rast} - I_0^\mathrm{GT} \Vert_2^2$ being the image loss for the initial frame only and $N_\mathrm{p}$ being the number of pixels in the images.
The texture regularization $\mathcal{R}_T = \mathrm{FD}(T)$ uses a simple finite difference formula (see supplementary material) to smooth the texture and interpolate unused texels with a small weight of $\lambda_T = 10^{-4}$.
We observe that keeping this texture fixed throughout the deformation optimization works best since the cloth is dominantly diffuse even though the optical material and the environment lighting can change the observed colors slightly.

\subsubsection{Shape-from-Template Optimization}
\label{section:SfT_optimization}

The shape optimization uses all frames of the video sequence and the texture of the previous step to start from the given initial geometry and successively reconstruct later frames.
Similar to previous work \cite{Phi-SfT,stotko2024physics}, we start to optimize only the first three frames and regularly add another frame every $5$ epochs.
During this phase, we optimize the cloth parameters $Y, B, S$ and external forces $\vec{F}_n^i$.
The loss
\begin{equation}
    \mathcal{L}_\mathrm{SfT} = \mathcal{L}_{\mathrm{im}} + \lambda_\mathrm{sil} \mathcal{L}_{\mathrm{sil}} + \lambda_E \mathcal{R}_E + \lambda_F \mathcal{R}_F
\end{equation}
is composed of four parts, the RGB image loss, the silhouette loss, and our two novel regularization terms.
Image and silhouette loss are defined as the mean of single-frame loss terms.
We are using the squared $L_2$ norm on the images to obtain smoother gradients for optimizing the motion.
Our key contribution to the 3D reconstruction task is the introduction of two regularization terms $\mathcal{R}_E, \mathcal{R}_F$ that deal with the depth ambiguity when trying to match the target images.
Without regularization, the cloth is moved regardless of how meaningful the deformation may be.
The gradients of the image and silhouette loss with respect to the vertex positions will dominantly be perpendicular to the viewing direction of the camera which results in crumpling of the cloth to match the target image instead of changing the depth.
We will elaborate on this effect in \Cref{section:ablation_study}.

\paragraph{Energy Regularization}
We observe that the optimization reduces the stiffness parameters as much as possible to get full control without soft restrictions from the simulation.
This leads to unrealistic crumpling as the fabric does not create a sufficiently large restoring force to counteract.
Hence, we guide the optimization towards a smoother surface by leveraging the internal cloth energy terms of all $N_\mathrm{f}$ currently used frames in a regularization
\begin{equation}
    \mathcal{R}_E = \frac{1}{N_\mathrm{f}} \sum_{n=1}^{N_\mathrm{f}} E_{n,\mathrm{SG}(Y)} + E_{n,\mathrm{SG}(B)} + E_{n,\mathrm{SG}(S)}
\end{equation}
where $\mathrm{SG}$ is the stop-gradient operator applied to the stiffness parameters.
Note that we have to stop these gradients because we want to regularize the geometry only.
Otherwise, the stiffness parameters will be minimized again which is what we aim to prevent in the first place.
In the case of cloth, we naturally choose the deformation energy which could be used in a similar way in the context of other solid materials as well.
The regularization weight is set to $\lambda_E = 2$ which results in a loss value that is 1--2 orders of magnitude lower than the image loss $\mathcal{L}_{\mathrm{im}}$ but is still sufficient to prevent crumpling.

\paragraph{Force Regularization}
In addition to the observed crumpling behavior, we observed minimal displacement along the normalized viewing direction $\vec{d}_n^i$ (from the camera position towards vertex $\vec{x}_n^i$), underscoring the crucial issue of depth ambiguity inherent in monocular data.
The reason lies in the information loss in the projection step during rendering.
In the supplementary material we show that the gradients of mesh vertices obtained in the backward pass are always perpendicular to the viewing direction.
Further contributions from the physical simulation may alter these gradients in the remaining backward pass but this information loss is inevitable and causes the depth ambiguity problem.
The external forces are most affected parameters due to their direct and immediate influence on the vertex positions.
Thus, we introduce the regularization
\begin{equation}
    \mathcal{R}_F = \frac{1}{N_\mathrm{f} N_\mathrm{V}} \sum_{n=1}^{N_\mathrm{f}} \sum_{i=1}^{N_\mathrm{V}} \left( \vec{F}_n^i - \frac{\langle \vec{F}_n^i , \vec{d}_n^i \rangle}{\langle \vec{d}_n^i , \vec{d}_n^i \rangle} \, \vec{d}_n^i \right) 
\end{equation}
which penalizes external forces $\vec{F}^i_n$ for each vertex $i$ and timestep $n$ along the image direction such that motion along the viewing direction $\vec{d}_n^i$ is preferred for all $N_\mathrm{V}$ vertices.
The regularization weight is set to $\lambda_F = 2 \cdot 10^{-4}$, producing a magnitude comparable to that of the energy regularization.

Both regularization terms are not meant to constrain the fabric's movements.
They are supposed to gently reduce unfavorable motion (\eg crumpling), guide the optimization towards more regular deformations and are thus similar to standard $L_1$ or $L_2$ regularizations.
Hence, the regularizations only contribute a fraction to the total loss value.
The effect of both regularizations will be discussed in detail in the ablation study (\Cref{section:ablation_study}).

\subsubsection{Appearance Estimation}

\begin{figure}
   \centering
   \includegraphics[width=0.94\linewidth]{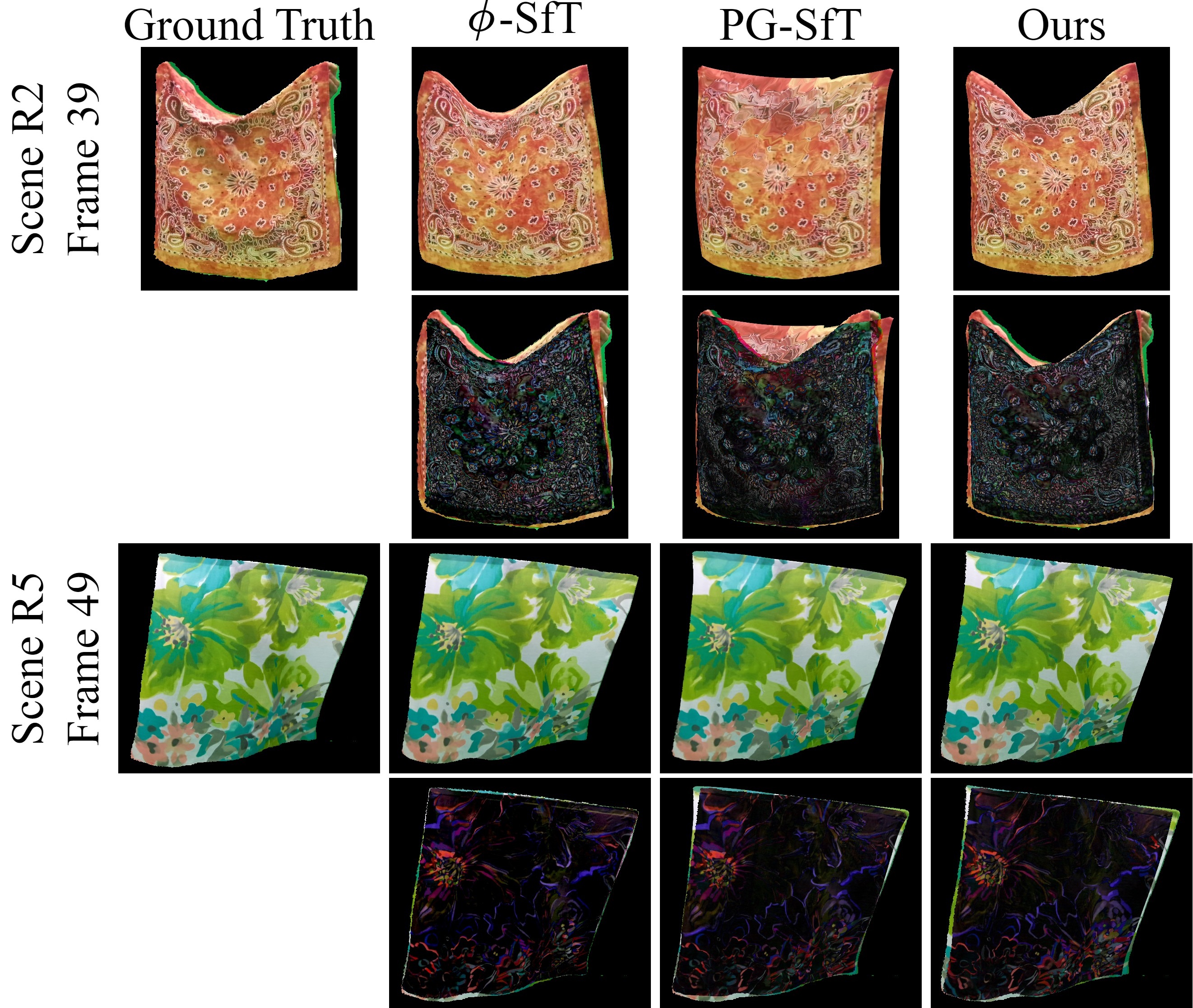}
   \caption{Qualitative comparison between our method and two competing methods for two example frames of different scenes.}
   \label{figure:qualitative_comparison}
\end{figure}

We are optimizing the decoder weights that generate the diffuse, roughness, and metallic textures, together with the environment map.
In order to separate the object color from the color in the lighting, we start with a gray environment map and only allow its brightness to change for the first $800$ epochs of this phase but not its hue.
Afterward, we slowly enable color information by using a linear interpolation between the mean of the color channels and the regular environment map.
This interpolation lasts for $400$ epochs after which we directly use the environment map with its RGB colors for another $800$ epochs.
Hence, we optimize the appearance and the illumination for $2000$ epochs in total.

In order to disambiguate the brightness of the appearance textures (especially the diffuse color) and the environment map, we slightly regularize the amplitude of the environment map $T_\mathrm{E}$ by applying a standard $L_1$ loss
with a weight of $\lambda_\mathrm{env} = 10^{-5}$.
Moreover, we apply a smoothness regularization $\mathcal{R}_\mathrm{s\_env} = \mathrm{FD}(T_\mathrm{E})$ to the environment map with a weight of $\lambda_\mathrm{s\_env} = 5 \cdot 10^{-5}$ similar to the texture smoothness regularization in \Cref{section:phase_texture_mapping}.
The full loss function for this phase is given by
\begin{equation}
    \mathcal{L}_\mathrm{Mat} = \mathcal{L}_{\mathrm{im}} + \lambda_\mathrm{env} \mathcal{R}_\mathrm{env} + \lambda_\mathrm{s\_env} \mathcal{R}_\mathrm{s\_env}.
\end{equation}

\begin{table*}
    \centering
    \setlength{\tabcolsep}{5.8pt}
    \begin{tabular}{l|ccccccccc|c}
        \toprule
        Method & R1 & R2 & R3 & R4 & R5 & R6 & R7 & R8 & R9 & Mean \\
        \midrule
        $\phi$-SfT \cite{Phi-SfT} & $6.68$ & $11.50$ & $\phantom{0}6.63$ & $11.10$ & $17.66$ & $9.30$ & $9.14$ & $3.93$ & $3.29$ & $8.80$ \\
        PG-SfT \cite{stotko2024physics} & $6.07$ & $4.06$ & $10.54$ & $13.32$ & $11.96$ & $15.29$ & $7.18$ & $9.64$ & $9.35$ & $9.71$ \\
        Ours & $\mathbf{1.01}$ & $\mathbf{1.55}$ & $\mathbf{2.71}$ & $\mathbf{8.29}$ & $\mathbf{7.12}$ & $\mathbf{2.18}$ & $\mathbf{4.84}$ & $\mathbf{1.23}$ & $\mathbf{1.04}$ & $\mathbf{3.33}$ \\
        \bottomrule
    \end{tabular}
    \caption{Quantitative comparison using the $L_2$ Chamfer distance $[\times 10^{-4} \, \mathrm{m}^2]$.}
    \label{table:quantitative_comparison_chamfer}
\end{table*}

\begin{table*}
    \centering
    \setlength{\tabcolsep}{5.8pt}
    \begin{tabular}{l|ccccccccc|c}
        \toprule
        Method & R1 & R2 & R3 & R4 & R5 & R6 & R7 & R8 & R9 & Mean \\
        \midrule
        $\phi$-SfT \cite{Phi-SfT} & $1242$ & $1406$ & $1032$ & $1231$ & $836$ & $1107$ & $1064$ & $1101$ & $923$ & $1105$ \\
        PG-SfT \cite{stotko2024physics} & $\mathbf{2.23}$ & $\mathbf{2.00}$ & $\mathbf{1.95}$ & $\mathbf{1.89}$ & $\mathbf{2.03}$ & $\mathbf{2.04}$ & $\mathbf{1.98}$ & $\mathbf{1.96}$ & $\mathbf{1.90}$ & $\mathbf{2.00}$ \\
        Ours & $34.1$ & $25.2$ & $29.6$ & $22.7$ & $49.0$ & $24.8$ & $29.5$ & $26.2$ & $28.5$ & $29.9$ \\
        \bottomrule
    \end{tabular}
    \caption{Runtime comparison for the geometry reconstruction $[\mathrm{min}]$.}
    \label{table:quantitative_comparison_runtime}
\end{table*}

\section{Evaluation}

We evaluate our approach on all real-world scenes of the $\phi$-SfT dataset \cite{Phi-SfT} due to its variety in motion and cloth sizes.
The dataset provides RGB images, depth images and point clouds for all frames but only the RGB images are used for the reconstruction task.
The videos were recorded in a capture stage with a green background and evenly distributed light sources.
First, we evaluate the 3D reconstruction qualitatively and quantitatively.
We compare our method to the reference implementations of $\phi$-SfT \cite{Phi-SfT} and PG-SfT \cite{stotko2024physics}.
Furthermore, we demonstrate the importance of our novel regularization terms in an ablation study (\Cref{section:ablation_study}).
Afterward, a qualitative evaluation of the appearance estimation is performed for the real-world scenes in \Cref{section:real_world_evaluation} and for synthetic scenes in \Cref{section:synthetic_evaluation}.
At the end, we elaborate on the limitations of our method.

\subsection{Motion Reconstruction}

\subsubsection{Qualitative Evaluation}

We compare our method to $\phi$-SfT~\cite{Phi-SfT} and PG-SfT~\cite{stotko2024physics} visually by presenting their reconstructions in late frames of scenes R2 and R5 that include significant deformations.
The reconstructions, ground truth, and difference images for these scenes are depicted in \Cref{figure:qualitative_comparison} and we present a full comparison of all scenes in a supplementary video.
PG-SfT produces errors in the textures and misses to follow the sharp folds, \eg. in scenes R2.
Both $\phi$-SfT and our method yield comparable visual quality with only small mismatches in all scenes.
However, note that our optimization finds a balance between a low image loss and a plausible geometry due to the regularization terms.

\begin{figure}
   \centering
   \includegraphics[width=0.99\linewidth]{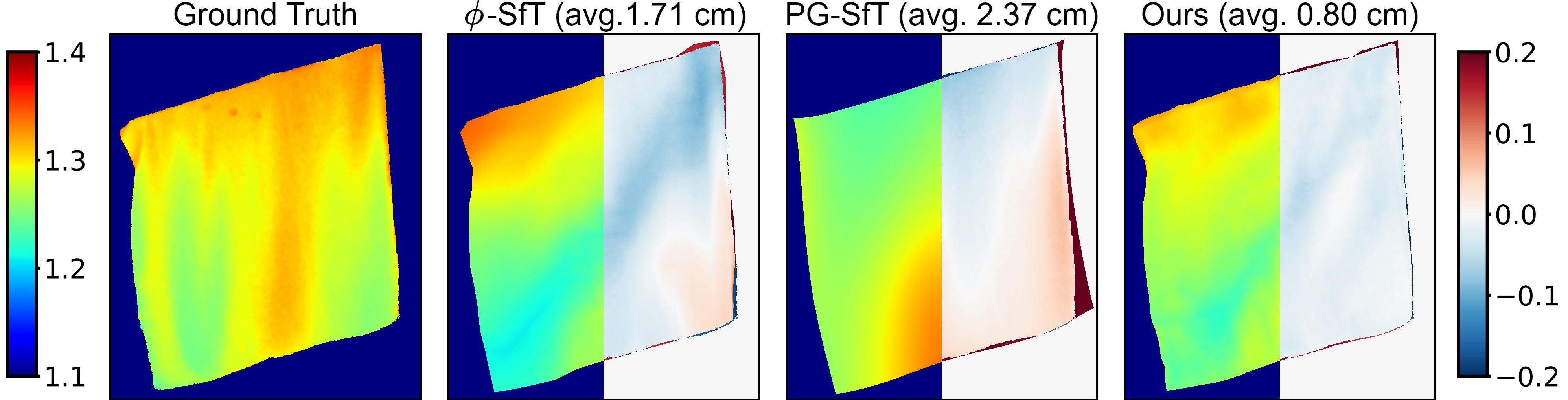}
   \caption{Depth images and differences for scene R6 frame 39.
   The average depth error during the whole video sequence is denoted in parenthesis.}
   \label{figure:depth_example}
\end{figure}

\subsubsection{Quantitative Evaluation}
\label{section:evaluation_motion_quantitative}

We quantitatively measure the reconstruction quality and runtime of our method and the ones we compare with.
The evaluations are performed on an Nvidia A100 GPU and an AMD Epyc 7713 CPU.
We refer to the supplemental PDF for a more detailed evaluation including more metrics, comparisons for all scenes, and all formulas.

\Cref{figure:depth_example} depicts depth images of all methods for the last frame in scene R6.
We clearly see less error in our depth map compared to the baseline methods, which is also confirmed by the average depth error in parenthesis.

\Cref{table:quantitative_comparison_chamfer} summarizes the $L_2$ Chamfer distances for all three methods.
Note that the results for $\phi$-SfT differ from the values reported in their paper because we do not perform any rigid alignment at the end~\cite{Phi-SfT}. 
Our methods reliably produce geometries with the lowest Chamfer distance often beating the other methods by a factor of at least two.
On average, we reduce the mean Chamfer distances of all scenes by $2.64 \times$ compared to $\phi$-SfT~\cite{Phi-SfT} and by $2.92 \times$ compared to PG-SfT~\cite{stotko2024physics}.
The table also demonstrates that the visual quality depicted in \Cref{figure:qualitative_comparison} does not give evidence about the similarity between the ground truth and the reconstructed geometry due to the missing depth perception in monocular views. 

We also evaluate the runtime of all methods.
\Cref{table:quantitative_comparison_runtime} lists the runtime to perform the geometry reconstruction (no subsequent appearance estimation for our method) in minutes for all scenes.
$\phi$-SfT is the slowest algorithm needing an average runtime of $18.4\,\mathrm{h}$ on the aforementioned hardware.
Our method performs the motion reconstruction after $29.9\,\mathrm{min}$ on average which is $36$ times faster than $\phi$-SfT.
Our runtime is dominantly affected by the number of frames in the scene resulting in a different number of optimization epochs.
The fastest method is PG-SfT with a runtime of only $2\,\mathrm{min}$ per scene with some slight variations due to their fast neural physics simulation~\cite{stotko2024physics}.
For the sake of completeness, we note that the appearance reconstruction part of our method takes additional $31.9\,\mathrm{min}$ on average to complete.
However, this runtime is independent of the 3D reconstruction time and would be the same when using the geometry of the other methods.

\subsubsection{Ablation Study}
\label{section:ablation_study}

\begin{table*}
    \centering
    \setlength{\tabcolsep}{5.8pt}
    \begin{tabular}{l|ccccccccc|c}
        \toprule
        Ablation & R1 & R2 & R3 & R4 & R5 & R6 & R7 & R8 & R9 & Mean \\
        \midrule
        w/o silhouette loss $\mathcal{L}_\mathrm{sil}$ & $4.36$ & $2.33$ & $2.24$ & $\mathbf{6.97}$ & $9.53$ & $\mathbf{1.60}$ & $5.95$ & $1.58$ & $1.10$ & $3.96$ \\
        w/o force regularization $\mathcal{R}_F$ & $7.69$ & $1.39$ & $2.51$ & $17.63$ & $10.45$ & $7.01$ & $\mathbf{3.41}$ & $2.06$ & $1.47$ & $5.96$ \\
        w/o energy regularization $\mathcal{R}_E$ & $2.54$ & $1.47$ & $\mathbf{1.91}$ & $28.84$ & $128.15$ & $3.68$ & $37.09$ & $3.66$ & $4.96$ & $23.59$ \\
        w/o both regularizations & $5.02$ & $\mathbf{1.13}$ & $3.11$ & $27.83$ & $120.92$ & $2.07$ & $37.83$ & $1.82$ & $2.89$ & $22.51$ \\
        Full model & $\mathbf{1.01}$ & $1.55$ & $2.70$ & $8.29$ & $\mathbf{7.12}$ & $2.18$ & $4.84$ & $\mathbf{1.23}$ & $\mathbf{1.04}$ & $\mathbf{3.33}$ \\
        \bottomrule
    \end{tabular}
    \caption{$L_2$ Chamfer distances of ablated models compared to the full model $[\times 10^{-4} \, \mathrm{m}^2]$.}
    \label{table:ablation_study}
\end{table*}

\begin{figure}
   \centering
   \includegraphics[width=0.85\linewidth]{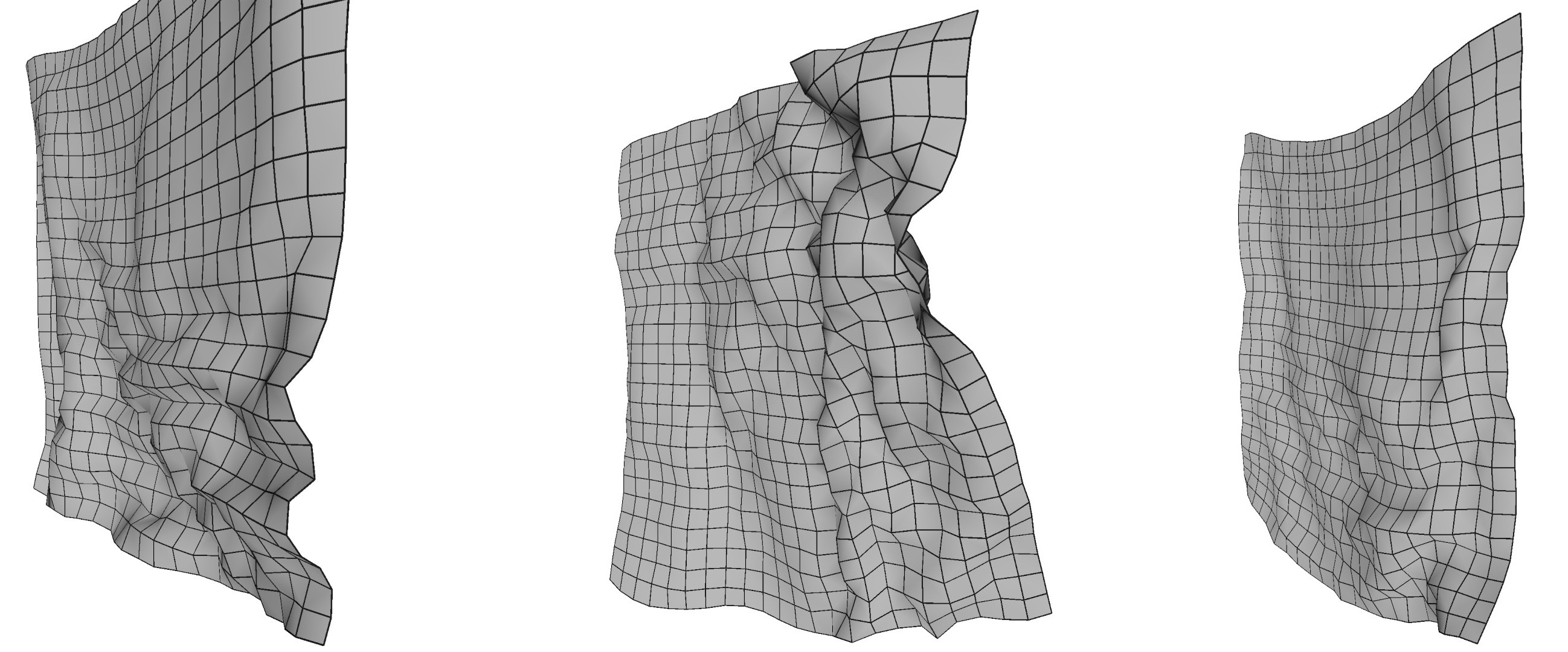}
   \caption{Crumpled fabric in scenes R4, R5, and R7 when using no regularization terms during the optimization.}
   \label{figure:ablation_crumpled}
\end{figure}

We perform an ablation study by disabling some parts in our reconstruction pipeline.
In particular, we evaluate the Chamfer distances of all scenes without using the silhouette loss $\mathcal{L}_\mathrm{sil}$, the force regularization $\mathcal{R}_F$, the energy regularization $\mathcal{R}_E$, or both regularizations.
The results are summarized in \Cref{table:ablation_study} and we will briefly elaborate on them.

The absence of the silhouette loss $\mathcal{L}_\mathrm{sil}$ has minimal impact on the overall reconstruction quality.
In some scenes, the quality improved when omitting the silhouette loss.
A simple explanation comes from the RGB images, where the background is already removed.
Thus, the mask information is already present in the RGB image loss.

Leaving out the force regularization $\mathcal{R}_F$ sometimes (\eg in scenes R2 and R3) does not change the quality significantly while other scenes (R1, R4, and R6) are heavily affected.
Especially the deformations in scenes R1 and R4 are induced by wind blowing approximately in the viewing direction of the camera.
These results nicely show the imbalance between gradient information in and orthogonal to the viewing direction caused by depth ambiguity in monocular video.
The largest variation in quality is present when disabling the energy regularization $\mathcal{R}_E$.
The Chamfer distances of three scenes are massively increased and dominate the mean Chamfer distance of all scenes.
Similar results are present when disabling both regularization losses.
As already discussed in \Cref{section:SfT_optimization}, the energy regularization prevents the optimization to crumple the cloth instead of moving it in a regular fashion.
We present examples of crumpled cloth geometry in \Cref{figure:ablation_crumpled} for the extreme cases in scenes R4, R5, and R7.
The only two scenes that do not change significantly or even improve in quality are R2 and R3.
Both of them show severe deformations that include a large fold which is slightly smoothed by the regularizations.

Our regularization terms are specifically designed to match the needs of our method and especially the physical simulation model.
$\phi$-SfT \cite{Phi-SfT} uses a coarser mesh and a more complex but slower physical model, which may be more resistant to crumpling and other effects.
Hence, we can not expect our regularization terms to improve its results in the same way.
PG-SfT \cite{stotko2024physics} already produces overly smooth reconstructions meaning that the regularizations will not be beneficial at all.


\subsection{Appearance Reconstruction}
\label{section:evaluation_appearance}

\subsubsection{Real-world Evaluation}
\label{section:real_world_evaluation}

\begin{figure}
   \centering
   \includegraphics[width=0.99\linewidth]{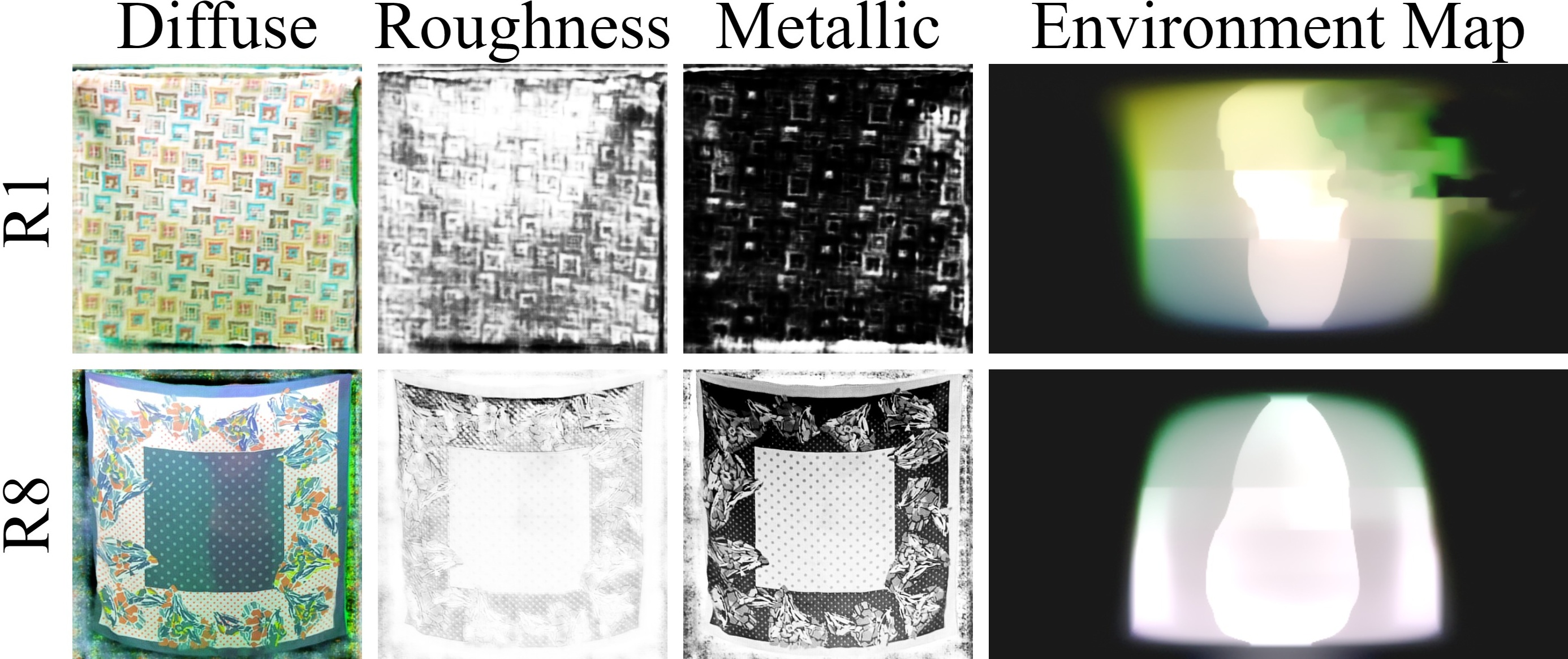}
   \caption{Two examples of estimated appearance of real-world scenes that are reconstructed only from one monocular view.}
   \label{figure:appearance_real}
\end{figure}

The $\phi$-SfT dataset does not contain any information about the fabric appearance.
We only know that the scenes were captured in a light stage with multiple light sources and green screen background~\cite{Phi-SfT}.
To the best of our knowledge, there is no method that reconstructs the SVBRDF using comparably low amount of information: The 3D reconstructed geometry of a deformable object captured by a monocular RGB camera without depth information.
Therefore, we report and discuss our results for the real scenes of the dataset and evaluate our method on scenes with synthetic appearance parameters.
Moreover, the supplemental material shows the lack of details when optimizing the appearance using the geometry reconstructed by PG-SfT.

In \Cref{figure:appearance_real}, the estimated SVBRDFs and environment maps of two example scenes (R1 and R8) are depicted.
The results of all scenes are summarized in the supplemental material and the corresponding renderings are given in the supplemental video.
The diffuse textures of all scenes look very similar to the images in the target video and resolve sharp details.
The roughness is quite high as expected for a diffuse-looking fabric but includes some spatial variations.
The metallic textures show a variety of different outcomes which sometimes depend on the pattern of the fabric and again include spatial variations even within similar regions.
Inside the environment map, only a large region in the center contains light information because only the light of that approximate hemisphere reflects off of the fabric and reaches the camera while the remaining parts approach zero due to the amplitude regularization.
Furthermore, the dominantly diffuse fabric only makes it possible to recover very blurry illumination conditions.
The environment map and diffuse color in scenes R1 and R8 are separated well and even a green border is visible.
Such a dominantly green hue in the environment map is also visible in various other scenes which likely represents the green screen background.

\subsubsection{Synthetic Evaluation}
\label{section:synthetic_evaluation}

\begin{figure}
   \centering
   \includegraphics[width=0.99\linewidth]{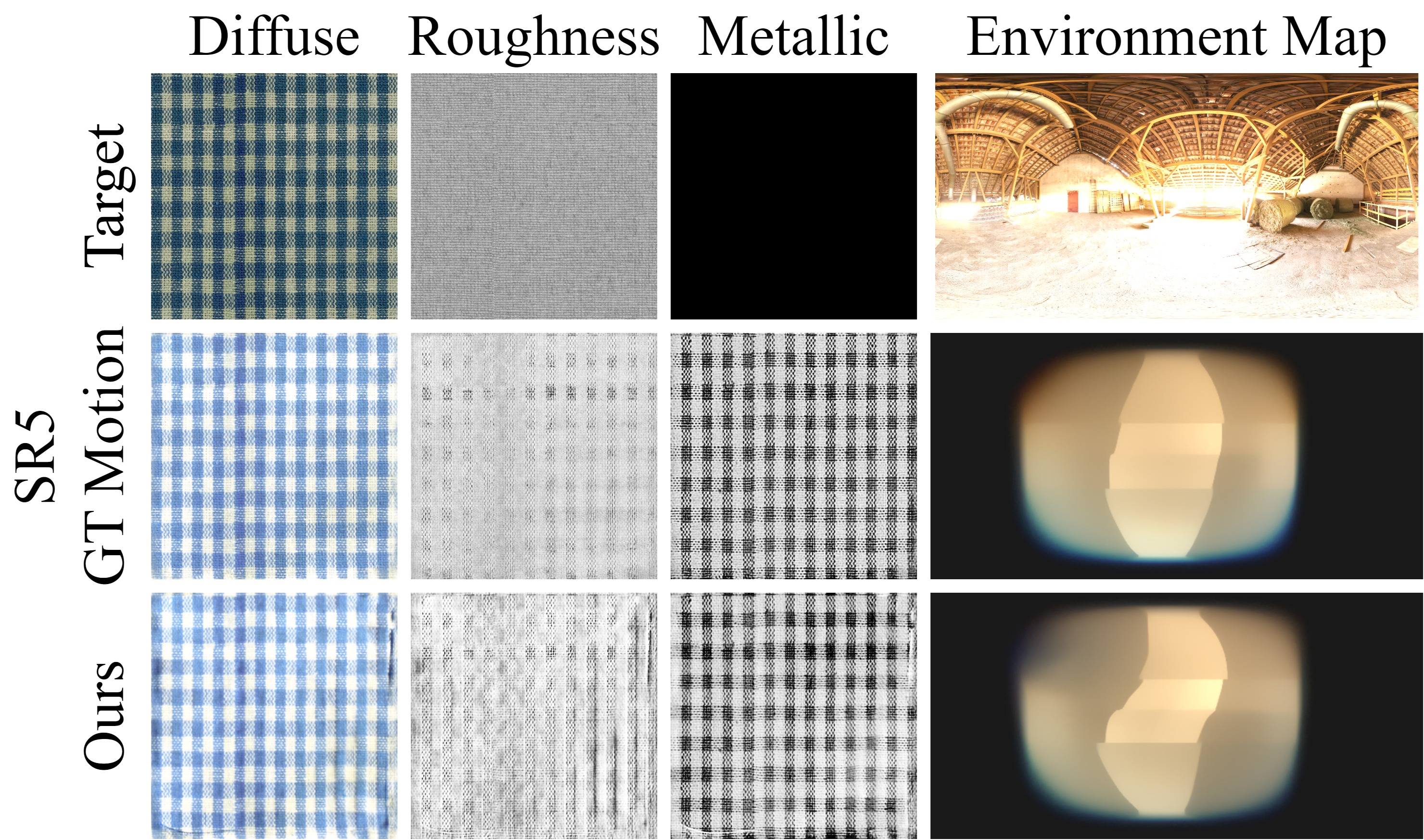}
   \caption{One example of estimating the appearance of the synthesized scene SR5.
   The first row shows the target textures used to render the images.
   The second and third rows display the estimated textures when employing the ground truth motion during the reconstruction or our reconstructed motion respectively.}
   \label{figure:appearance_synthetic}
\end{figure}

To conduct a more detailed evaluation of the appearance estimation, five synthetic scenes were created, each with renderings of different appearance parameters\footnote{All textures were taken from PolyHaven \url{https://polyhaven.com/}}.
To ensure the highest degree of similarity with the $\phi$-SfT dataset, we reuse the reconstructed motion of our algorithm and save renderings with novel appearance parameters.
We call these scene SR1 to SR5 ("synthesized RX") with the motion taken from R1 to R5, respectively.

The results of reconstructing scene SR5 are given in \Cref{figure:appearance_synthetic} while all remaining synthesized scenes are given in the supplemental PDF.
The first row in the figure displays the ground truth appearance textures and the environment map used for creating the rendered target sequence.
Since we know the data that generated the synthesized scenes, we are able to perform the appearance estimation using the ground truth motion and report the results in the second row.
Hence, we get perfect correspondences for mapping the texture onto the geometry resulting in the appearance estimation possible with our method.
We can see that despite some patterns in the roughness and metallic map we get reasonable results for the whole appearance including high-resolution details.
Even the brown-colored light is correctly separated into the environment map.
The last row depicts the optimized textures when using a reconstructed geometry similar to reconstructing the real scenes.
Especially the scenes with non-metallic materials (\eg the one given in \Cref{figure:appearance_synthetic}) yield almost the same results as for the ground truth motion.

\subsection{Limitations}

\subsubsection{3D reconstruction}

During our first phase, we perform a projective texture mapping that provides the main information for the 3D reconstruction.
This approach does not reproduce specular highlights from shiny material that might be present when capturing \eg silk fabric.
However, estimating the appearance parameters reliably from a monocular view is a challenging task in itself, without knowing the deformation.
One possibility could be to jointly reconstruct the motion and the SVBRDF either simultaneously or in an alternating fashion.

Furthermore, despite the high quality achieved in comparison to existing approaches, \Cref{figure:qualitative_comparison} still shows discrepancies between our reconstructed motion and the target sequence.
In this regard, we also showed that our ablated method yield higher quality in specific scenes.
Thus, there is the potential to further improve on the quality by finding a universally accurate model for all scenes at the same time.

\subsubsection{Appearance Estimation}

The SVBRDF optimization is mainly limited by the amount of information available during the optimization.
However, there is still room to improve the current results since in some cases the diffuse texture includes inpaintings or the environment map includes some diffuse color.
Moreover, the roughness and metallic maps are inpainted as well with regions similar to the diffuse texture pattern.

\section{Conclusion}

We have presented a Shape-from-Template method to reconstruct the dynamic geometry of deforming fabric employing a physically simulated model.
Two novel regularization terms have been introduced, which improve the reconstruction quality by guiding toward a distinct solution to the depth ambiguity problem in monocular videos.
As a result, we are able to reduce the chamfer distance between the ground truth and the reconstructed geometry by factors of $2.64$ and $2.92$ compared to the state-of-the-art methods $\phi$-SfT and PG-SfT respectively while requiring moderate runtime.
Additionally, we show that it is possible to conduct a material estimation only using the precisely reconstructed geometry and the monocular RGB video data.
We have demonstrated that even this limited information is sufficient to obtain detailed appearance textures.
We hope that our approach will encourage future work in this domain towards the goal of reducing the overall assumptions for material estimation.

\section*{Acknowledgements}

This work has been funded by the DFG project KL 1142/11-2 (DFG Research Unit FOR 2535 Anticipating Human Behavior), by the Federal Ministry of Education and Research of Germany and the state of North-Rhine Westphalia as part of the Lamarr-Institute for Machine Learning and Artificial Intelligence and the InVirtuo 4.0 project, and additionally by the Federal Ministry of Education and Research under grant no. 01IS22094A WEST-AI.
{
    \small
    \bibliographystyle{ieeenat_fullname}
    \bibliography{main}
}

\begin{figure*}[t]
\Large
\centering
\textbf{Supplementary Material}
\end{figure*}

\noindent
This supplementary material contains further information about the physical simulation and the rendering gradients in \Cref{section:method}.
Further implementation details and additional formulas are summarized in \Cref{section:implementation_details}.
The complete evaluation of the appearance estimation as one possible downstream task for the 3D reconstruction is given in \Cref{section:evaluation}.

\section{Method}
\label{section:method}

\subsection{Numerical Integration Scheme}

For simulation, we discretize the time into constant steps of $\Delta t = 5 \, \mathrm{ms}$ resulting in all time-dependent quantities to be discretized accordingly, \eg vertex positions $\vec{x}_n = \vec{x}(n \Delta t)$.
The update rules when applying the backward Euler schemes to Newton's law are given by
\begin{align}
    \vec{v}_{n+1} & = \vec{v}_n + \Delta t \vec{a}_{n+1} = \vec{v}_n + \Delta t M^{-1} \vec{F}_{n+1} \label{v_update} \\
    \vec{x}_{n+1} & = \vec{x}_n + \Delta t \vec{v}_{n+1}. \label{x_update}
\end{align}
However, these rules require information about the time step that we are trying to compute in the first place.
This problem can be solved by introducing the differences $\Delta v = \vec{v}_{n+1} - \vec{v}_n$ and $\Delta x = \vec{x}_{n+1} - \vec{x}_n = \Delta t \vec{v}_{n+1}$ and approximating the force by a first-order Taylor series \cite{baraff_witkin_large_steps}:
\begin{equation}
    \begin{split}
        &\vec{F}_{n+1} = \vec{F}(\vec{x}_n + \Delta \vec{x}, \vec{v}_n + \Delta \vec{v}) \\
        &\approx \vec{F}(\vec{x}_n, \vec{v}_n) + \frac{\partial \vec{F}(\vec{x}_n, \vec{v}_n)}{\partial \vec{x}} \Delta \vec{x} + \frac{\partial \vec{F}(\vec{x}_n, \vec{v}_n)}{\partial \vec{v}} \Delta \vec{v} \\
        &= \vec{F}_n + \Delta t  \frac{\partial \vec{F}_n}{\partial \vec{x}} \vec{v}_{n+1} + \frac{\partial \vec{F}_n}{\partial \vec{v}} \Delta \vec{v}
    \end{split}
\end{equation}
Inserting this approximation into \Cref{v_update} yields the expression
\begin{equation}
    M \vec{v}_{n+1} = M \vec{v}_n + \Delta t \vec{F}_n + (\Delta t)^2 \frac{\partial \vec{F}_n}{\partial \vec{x}} \vec{v}_{n+1} + \Delta t \frac{\partial \vec{F}_n}{\partial \vec{v}} \Delta \vec{v}
\end{equation}
which can be transformed to get
\begin{multline}
\label{equation:euler_update}
    \left( M - (\Delta t)^2 \frac{\partial \vec{F}_n}{\partial \vec{x}} - \Delta t \frac{\partial \vec{F}_n}{\partial \vec{v}} \right) \vec{v}_{n+1} = \\ M \vec{v}_n + \Delta t \vec{F}_n + \Delta t \frac{\partial \vec{F}_n}{\partial \vec{v}} \vec{v}_n
\end{multline}
for the velocity update.
This linear system of equations can be solved using only quantities from the current time step.

\subsection{Energy Terms}
\label{subsection:energies}

The forces acting on the cloth can be split into internal forces (stretching, bending, shearing) and external forces (gravity, wind, and other contributions).
We model the internal forces explicitly by imposing corresponding deformation energies and obtaining the forces using the derivatives $\vec{F} = - \frac{\partial E}{\partial \vec{x}}$.
The following paragraphs will describe each contribution in more detail.

\paragraph{Stretching}
A stretching energy limits the individual changes in length for edges $\vec{e}^{\,ij}$ between two connected vertices $i$ and $j$.
We impose a Hookean energy term
\begin{equation}
    E_Y = \frac{1}{2} Y \left( \Vert \vec{e}^{\,ij} \Vert - L_0^{ij} \right)^2
\end{equation}
that penalizes deviations from the rest length $L_0^{ij}$.
The parameter $Y$ is called stretching stiffness or Young's modulus and scales the amount of stress required to create a certain strain.

\begin{figure}
    \centering
    \includegraphics[width=0.99\linewidth]{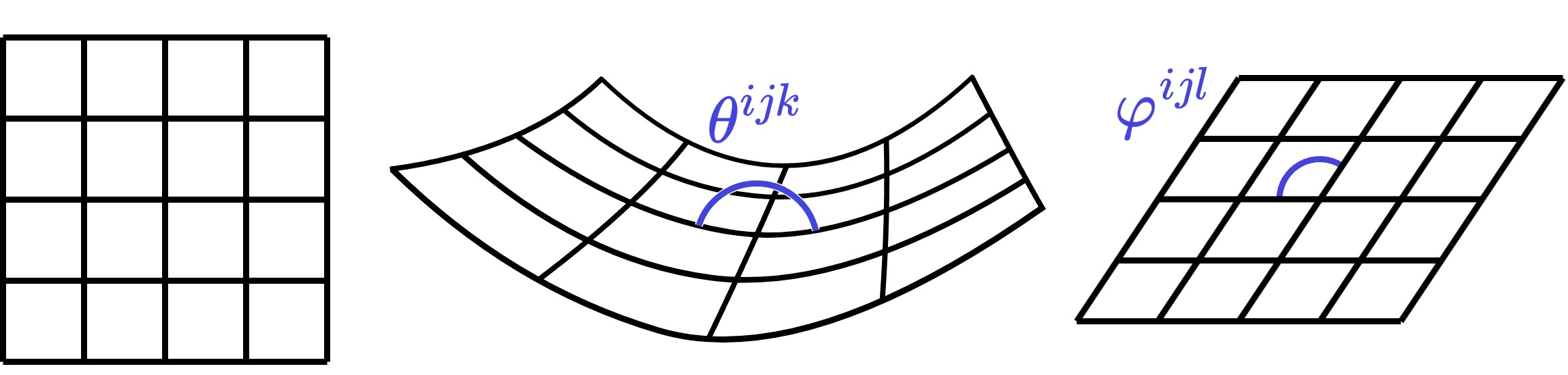}
    \caption{Schematic view of a fabric in its rest state, a bend fabric, and a sheared one.}
    \label{figure:deformations}
\end{figure}

\paragraph{Bending}
We also define two resistances against the deformation of angles within the geometry.
The bending energy describes the resistance against a curvature relative to the rest state.
In the case of a quad mesh, we use pairs~$(\vec{e}^{\,ij}, \vec{e}^{\,jk})$ of connected edges that form a straight path on the geometry (see \Cref{figure:deformations}).
For an enclosed angle $\theta^{ijk} = \sphericalangle(\vec{e}^{\,ij}, \vec{e}^{\,jk})$ we define the bending energy
\begin{equation}
    E_B = \frac{1}{2} B \left( \theta^{ijk} - \theta_0^{ijk} \right)^2
\end{equation}
with the rest angle $\theta_0^{ijk}$ and a bending stiffness $B$ \cite{Curvature_Sullivan2008,Cirio_yarn_1}.

\paragraph{Shearing}
Furthermore, an energy term is required that penalizes shearing deformations of the geometry, as these do not necessarily result in stretching or bending of the geometry.
In real cloth, such deformations will push different yarns together, resulting in a repulsive force from the collisions.
An energy term can again be modeled using the angles $\varphi^{ijl} = \sphericalangle(\vec{e}^{\,ij}, \vec{e}^{\,jl})$ of all remaining pairs of connected edges that were not used for bending (\Cref{figure:deformations}).
The functional form of the shearing energy
\begin{equation}
    E_S = \frac{1}{2} S \left( \varphi^{ijl} - \varphi_0^{ijl} \right)^2
\end{equation}
is the same as for the bending energy but contains a different scaling parameter $S$, the shearing stiffness.
It would also easily be possible to allow for independent stiffness parameters for each edge or pair of edges.
However, for the sake of simplicity, we opted to use a homogeneous cloth model.

\paragraph{External Forces}
All other forces acting on the cloth are summarized in external forces $\vec{F}_\mathrm{ext}$.
In general, these forces are not conservative and, thus, can not be modeled as the gradient of an energy term.
Therefore we treat them as arbitrary vectors $\vec{F}_n^i$ for each vertex $i = 1,\dots,N_\mathrm{V}$ at each frame $n = 1,\dots,N_\mathrm{f}$.
Two important examples in our scenario are the presence of wind and manipulations by people.
Due to their many degrees of freedom, the external forces are able to model the intricate internal cloth forces as well.
However, previous work has demonstrated that this approach is much more challenging to optimize and often results in inferior reconstructions~\cite{Phi-SfT}.
We can observe similar effects, as the forces begin to crumple the fabric together when our regularization terms are omitted, see the ablation study of the main paper (Section 4.1.3).
For more control we split the external forces $\vec{F}_n^i = \vec{C} + \vec{D}_n^i$ into a spatially and temporally constant part~$\vec{C}$ and a dynamic part~$\vec{D}_n^i$.

\subsection{Simulation Details}

\paragraph{Time Discretization}
We perform a physical simulation using four constant steps of $\Delta t = 5 \, \mathrm{ms}$ between frames to match the $20 \, \mathrm{ms}$ video data.
For the sake of simplicity, we keep the external forces constant for time steps in between frames, \ie we reuse the same external forces $\vec{F}_n^i$ four times until we switch to the next forces $\vec{F}_{n+1}^i$.
Hence, there is no difference between simulation steps and frames other than the simulation error and stability.
Individual forces per simulation time step are easy to implement but may require some smoothness regularization to get reasonable interpolation steps due to missing target frames to compare to.

\paragraph{Mass}
In order to apply \Cref{equation:euler_update}, we need a mass matrix $M = m I$ associated to each vertex of the cloth.
To do so, we define an area density coefficient $\rho$.
We triangulate our template mesh, calculate the mass of each triangle face, and distribute a third of the face mass to each of its vertices.
This leads to a relatively homogeneous mass distribution inside the mesh and lower masses at the mesh border.
Note that the cloth motion is determined by the acceleration $\vec{a} = M^{-1} \vec{F}$ only and scaling force as well as the mass by the same amount will result in the same motion.
Hence, we keep the area density and therefore the masses constant using $\rho = 0.1 \, \frac{\mathrm{kg}}{\mathrm{m}^2}$ while optimizing the forces.

\paragraph{Damping}
We also include an explicit damping term due to the large air resistance present for cloth.
We consider the minimalistic damping model
\begin{equation}
    \vec{v}_{n} = \delta \, \vec{v}'_{n}
\end{equation}
that scales down the vertex velocities by a factor of $\delta$ after the update steps in \Cref{equation:euler_update,x_update} are applied.
The damping is kept constant at $\delta = 0.9$ as well, since small changes will affect the whole movement and make optimization difficult.

\subsection{Orthogonality of Gradients}

\setlength{\columnsep}{7pt}
\begin{wrapfigure}[11]{R}{0.48\linewidth}
    \centering
    \vspace{-3mm}
    \includegraphics[width=\linewidth]{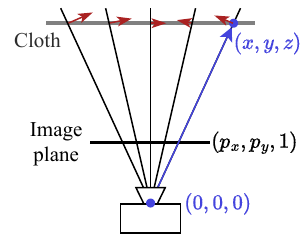}
    \caption{Projection step in camera coordinates.}
    \label{figure:gradients_proof}
\end{wrapfigure}

The gradient of the loss function with respect to some arbitrary (physical) parameter~$q$ can be decomposed using the chain-rule.
We assume the parameter $q$ to have some influence on a vertex at position $(x \enspace y \enspace z)$ that will be projected on pixel $(p_x \enspace p_y)$ during rendering.
An illustration of the projection step is depicted in \Cref{figure:gradients_proof}.
We now decompose the gradient
\begin{equation}
\label{equation:full_chain_rule}
    \pdv{\mathcal{L}}{q} = \pdv{\mathcal{L}}{(p_x \enspace p_y)} \pdv{(p_x \enspace p_y)}{(x \enspace y \enspace z)} \pdv{(x \enspace y \enspace z)}{q}
\end{equation}
into three parts denoting contributions for the projection step itself as well as the calculations before and afterwards.
Our renderer makes use of the OpenCV camera convention such that the camera intrinsic parameters $c_x, c_y, f_x, f_y$ determine on which pixel a vertex will be projected:
\begin{equation}
    \begin{pmatrix}
        p_x \\
        p_y
    \end{pmatrix}
    =
    \begin{pmatrix}
        \frac{f_x x}{z} + c_x \\
        \frac{f_y y}{z} + c_y
    \end{pmatrix}
\end{equation}
Its derivative with respect to the vertex coordinates is therefore
\begin{equation}
    \pdv{(p_x \enspace p_y)}{(x \enspace y \enspace z)} =
    \begin{pmatrix}
        \frac{f_x}{z} & 0 & - \frac{f_x x}{z^2} \\
        0 & \frac{f_y}{z} & - \frac{f_y y}{z^2}
    \end{pmatrix}
\end{equation}
Similar calculations can be done for other camera conventions as well.
An arbitrary image gradient $\pdv{\mathcal{L}}{(p_x \enspace p_y)} = (a \enspace b)$ now yields the following $3$D gradient for the vertex position:
\begin{align}
    \pdv{\mathcal{L}}{(x \enspace y \enspace z)} & =
    \begin{pmatrix}
        a & b
    \end{pmatrix}
    \begin{pmatrix}
        \frac{f_x}{z} & 0 & - \frac{f_x x}{z^2} \\
        0 & \frac{f_y}{z} & - \frac{f_y y}{z^2}
    \end{pmatrix}
    \\
    & \hspace{-3mm} = a
    \begin{pmatrix}
        \frac{f_x}{z} & 0 & - \frac{f_x x}{z^2}
    \end{pmatrix}
    + b
    \begin{pmatrix}
        0 & \frac{f_y}{z} & - \frac{f_y y}{z^2}
    \end{pmatrix}
    \\
    & \hspace{-3mm} =
    \begin{pmatrix}
        a \frac{f_x}{z} & b \frac{f_y}{z} & - a \frac{f_x x}{z^2} - b \frac{f_y y}{z^2}
    \end{pmatrix}
\end{align}
The (unnormalized) viewing direction towards the vertex in camera space is exactly $\vec{d} = (x \enspace y \enspace z)$ and the orthogonality follows directly:
\begin{align}
    \left \langle \pdv{\mathcal{L}}{(x \enspace y \enspace z)}, \vec{d} \right \rangle
    & =
    \begin{pmatrix}
        a \frac{f_x}{z} & b \frac{f_y}{z} & - a \frac{f_x x}{z^2} - b \frac{f_y y}{z^2}
    \end{pmatrix}
    \begin{pmatrix}
        x \\ y \\ z
    \end{pmatrix}
    \nonumber \\
    & = 0
\end{align}
As a consequence, the orthogonality of the final gradients $\pdv{\mathcal{L}}{q}$ can only be altered by calculations before the projection step happens.
In our case, the physical simulation introduces complex interaction between vertices such that the term $\pdv{(x \enspace y \enspace z)}{q}$ in \Cref{equation:full_chain_rule} breaks the perfect orthogonality.
Nonetheless, the loss of information during the projection causes the depth ambiguity problem and heavily influences all gradients with explicit implications for large portions of the gradient computation.

While this property of the rendering gradients is not new, our regularization terms are designed to utilize this knowledge and enhance the gradients in order to improve on $3$D reconstruction tasks in monocular setups.
While the regularizations are closely related to our specific physical simulation, the presented concepts are applicable in a variety of scenarios.

\begin{table}
    \centering
    \def\arraystretch{1.1}
    \resizebox{\columnwidth}{!}{%
    \begin{tabular}{l|cccc}
        \toprule
        Parameter & Init & Min & Max & lr \\
        \midrule
        $\mathrm{log}_{10}(Y)$ & $\mathrm{log}_{10}(200)$ & $1$ & $3$ & $0.02$ \\
        $\mathrm{log}_{10}(B)$ & $-3$ & $-4$ & $-2$ & $0.02$ \\
        $\mathrm{log}_{10}(S)$ & $-4$ & $-5$ & $-2$ & $0.02$ \\
        $\vec{C}$ & $(0, -1, 0)$ & - & - & $0.1$ \\
        $\vec{D}_n^i$ & $\vec{0}$ & - & - & $0.2$ \\
        Texture $T$ & $0.5$ & $0$ & $1$ & $0.05$ \\
        Diffuse $T_\mathrm{D}$ & $(0.5)$ & $0$ & $1$ & $2\cdot10^{-4}$ \\
        Roughness $T_\mathrm{R}$ & $(0.5)$ & $10^{-3}$ & $1$ & $2\cdot10^{-4}$ \\
        Metallic $T_\mathrm{M}$ & $(0.5)$ & $0$ & $1$ & $2\cdot10^{-4}$ \\
        Environment $T_\mathrm{E}$ & $1.5$ & $0$ & - & $0.01$ \\
        \bottomrule
    \end{tabular}
    }
    \caption{Initial values, limits, and learning rates of all optimized parameters.}
    \label{table:values_and_learning_rates}
\end{table}

\section{Implementation Details}
\label{section:implementation_details}

\subsection{Simulation Scheme}

As a consequence of the spatial discretization into discrete vertices, the energy terms dependent not only on a single vertex but also on neighboring vertices.
In order to get an accurate update step for all vertices at once, we have to include their influence as well.
Hence, the derivatives $\frac{\partial \vec{F}_n}{\partial \vec{x}}$ in \Cref{equation:euler_update} have to be computed for all vertices participating in the calculation.
All derivatives are computed automatically using the automatic differentiation abilities of PyTorch \cite{PyTorch} such that only the simple energy functions have to be implemented.

It is possible to infer the physical parameters $Y, B, S$ as well as the external forces acting on the cloth.
However, we are only able to infer the net forces that determine the motion of the cloth. 
Canceling contributions such as gravity and the counteracting holding forces can not be reconstructed individually.
Nonetheless, the reconstructed physical parameters and forces can be used to recreate or even change the given dynamics.

\subsection{Stable Gradients}

Some functions or derivatives introduce numerically instable or undesired expressions in the forward or backward pass of the optimization.
We mitigate this issue by limiting all such values to always maintain some distance to all mathematical singularities.
However, there may be dramatic variations in the gradients that remain, and we address this issue through two methods.
The first is to clip the gradients to a maximum length of $1000$.
The second is to employ an automatic gradient clipping algorithm \cite{autoclip} that clips gradients based on their length in previous epochs.

\subsection{Initial Values, Limits, and Learning Rates}
We always start with the same parameters to reconstruct the cloth.
Similarly, some parameters are limited to lie between a minimum and maximum value to ensure a stable simulation.
All these values are summarized in \Cref{table:values_and_learning_rates}.
Note, that we actually optimize the logarithm of the three stiffness parameters (when using SI units), \eg $\mathrm{log}_{10}(Y)$, due to their large range of reasonable values of which we allow at least two orders of magnitude.
With our hyperparameters the stiffness parameters do not reach their limits during optimization.
The different textures and the environment map are initialized with a uniform gray color.
In the case of the material textures (diffuse, roughness, and metallic textures), the decoder starts the optimization with gray textures.
Note, that the roughness values have to be strictly positive due to singularities at zero during the rendering.

\subsection{Comparison of all evaluated Methods}
\label{section:method_comparison}

All three methods use ground truth geometry in the first frame but remeshing introduces minor differences between all of them.
$\phi$-SfT~\cite{Phi-SfT} uses irregular meshes with approximately $400$ vertices per scene.
They employ arcsim~\cite{arcsim} to perform a sophisticated cloth simulation based on continuum mechanics and PyTorch3D~\cite{pytorch3D} for rendering, both of which having high computational costs.
PG-SfT~\cite{stotko2024physics} needs regular rectangular meshes with $32 \times 32 = 1024$ vertices but does not apply any texture mapping resulting in distorted textures compared to ground truth.
The physical simulation is performed by a fast physics-informed neural network that was trained in a self-supervised way.
Although the neural network computes the simulation results very quickly, its quality is limited to low-frequency motion due to its simple convolutional architectures.
The rendering is done using nvdiffrast~\cite{nvdiffrast} for fast and differentiable rendering.
Our approach combines the benefit of both predecessors: we employ a fast yet accurate simulation based on a mass-spring system.
It produces versatile results, is easy to extend and to implement.
For the reconstruction task we use regular meshes with $25 \times 25 = 625$ vertices but arbitrary meshes can be simulated as well.
We make use of nvdiffrast as well for fast rendering but also include nvdiffrecmc~\cite{nvdiffrecmc} for high-quality rendering based on ray tracing.

\subsection{Additional Formulas}
\label{section:formulas}

\subsubsection{Finite Differences}

Within the texture mapping phase, the texture $T$ is smoothed using the regularization $\mathcal{R}_T = \mathrm{FD}(T)$ with
\begin{align}
    \begin{split}
        \mathrm{FD}(T) & := \hspace{8pt} \frac{1}{(w-1) h} \sum_{i = 1}^{w - 1} \sum_{j = 1}^{h} \Vert T_{i+1,j} - T_{i,j} \Vert_1 \\
        & \hspace{14pt} + \frac{1}{w (h-1)} \sum_{i = 1}^{w} \sum_{j = 1}^{h - 1} \Vert T_{i,j+1} - T_{i,j} \Vert_1
    \end{split}
\end{align}
being the formula for the mean of all neighboring finite differences in the texture.
The same regularization is employed to smooth the environment map $T_\mathrm{E}$ during the appearance estimation phase.

\subsubsection{Depth Differences}
\label{section:depth_formula}

We are able to render depth images $D_\mathrm{R}$ of the reconstructed mesh and compare them with ground truth $D_\mathrm{GT}$.
We only consider pixels $i,j$ with depth values within the region of interest (ROI) in which both the ground truth as well as the reconstructed geometry yield valid depth values:
\begin{equation}
    \delta(D_\mathrm{GT}, D_\mathrm{R}) = \frac{1}{\vert \mathrm{ROI}_n \vert} \sum_{i, j \in \mathrm{ROI}} \Vert D_{ij,\mathrm{R}} - D_{ij,\mathrm{GT}} \Vert_2
\end{equation}
We report the average depth error for all frames in a scene as the final metric.

\subsubsection{Chamfer Distance}
\label{section:chamfer_formula}

\begin{figure}
   \centering
   \includegraphics[width=0.95\linewidth]{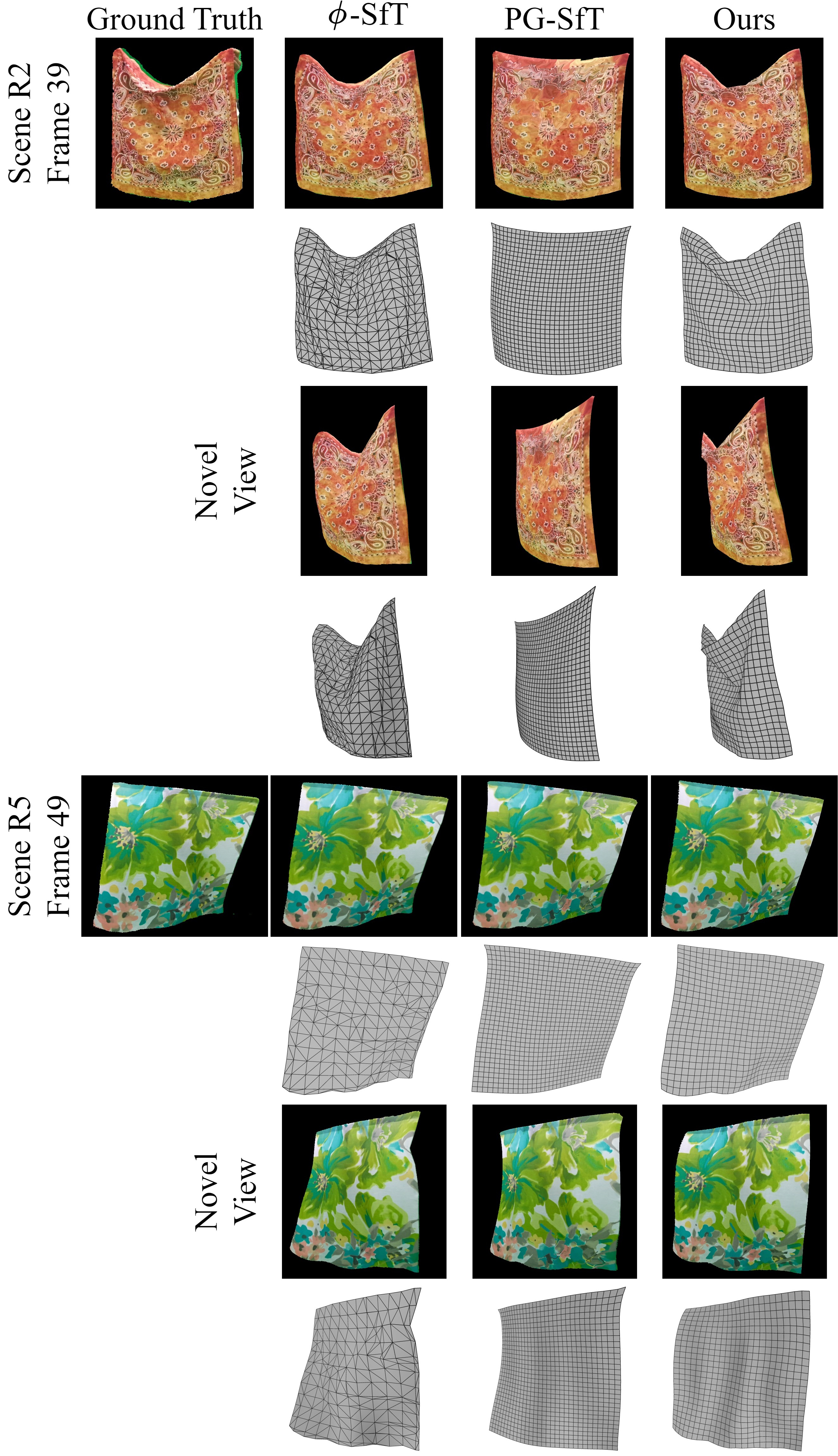}
   \caption{Renderings and mesh views of all methods for two different scenes.}
   \label{figure:qualitative_comparison_supplement}
\end{figure}

Similar to previous work \cite{Phi-SfT,stotko2024physics}, we use the symmetric Chamfer distance to quantify the quality of our 3D reconstructed geometry.
\begin{equation}
    \begin{split}
        \mathbf{CD}_p(P_\mathrm{GT}, P_\mathrm{R}) & = \hspace{12pt} \frac{1}{\vert P_\mathrm{R} \vert} \hspace{4pt} \sum_{\vec{r} \in P_\mathrm{R}} \min_{\vec{t} \in P_\mathrm{GT}} \Vert \vec{r} - \vec{t} \Vert_2^p \\
        & \hspace{10pt} + \frac{1}{\vert P_\mathrm{GT} \vert} \sum_{\vec{t} \in P_\mathrm{GT}} \min_{\vec{r} \in P_\mathrm{R}} \Vert \vec{r} - \vec{t} \Vert_2^p
    \end{split}
\end{equation}
The chamfer distance is computed by comparing a ground truth point cloud $P_\mathrm{GT}$ from a depth sensor with a sampled point cloud $P_\mathrm{R}$ on the reconstructed fabric mesh and averaging these values over all frames in a scene.
The point cloud of the reconstructed mesh $P_\mathrm{R}$ will always have the same number of points as the ground truth point cloud $P_\mathrm{GT}$.
We use the $L_1$ Chamfer distance $\mathbf{CD}_1$ as well as the $L_2$ Chamfer distance $\mathbf{CD}_2$ and compare both of them.
The use of linear and quadratic weighting facilitates the identification of outliers.

\subsubsection{Point to Surface Distance}
\label{section:p2s_formula}

We also report the symmetric point-to-surface (p2s) distance between the ground truth point cloud $P_\mathrm{GT}$ and the reconstructed triangulated surface $T_\mathrm{R}$
\begin{equation}
    \begin{split}
        \mathrm{p2s}_p(P_\mathrm{GT}, T_\mathrm{R}) & = \hspace{12pt} \frac{1}{\vert T_\mathrm{R} \vert} \hspace{4pt} \sum_{\vec{r} \in T_\mathrm{R}} \min_{\vec{t} \in P_\mathrm{GT}} \mathrm{p2t}(t, r)^p \\
        & \hspace{10pt} + \frac{1}{\vert P_\mathrm{GT} \vert} \sum_{\vec{t} \in P_\mathrm{GT}} \min_{\vec{r} \in P_\mathrm{R}} \mathrm{p2t}(t, r)^p
    \end{split}
\end{equation}
using the point-to-triangle distances $\mathrm{p2t}$.
Note that in our case the point cloud is the ground truth data and the mesh is the reconstructed surface.
Thus, a one-sided metric that averages the minimal distances from each point in the point cloud to the closest point of the mesh surface would not account for overly large reconstructed surfaces.
Again, we average the point-to-surface distances for all frames in a scene for the final metric.
Similar to the Chamfer distance, we report the $L_1$ and $L_2$ p2s distance and compare both of them.

\begin{figure}
   \centering
   \includegraphics[width=0.99\linewidth]{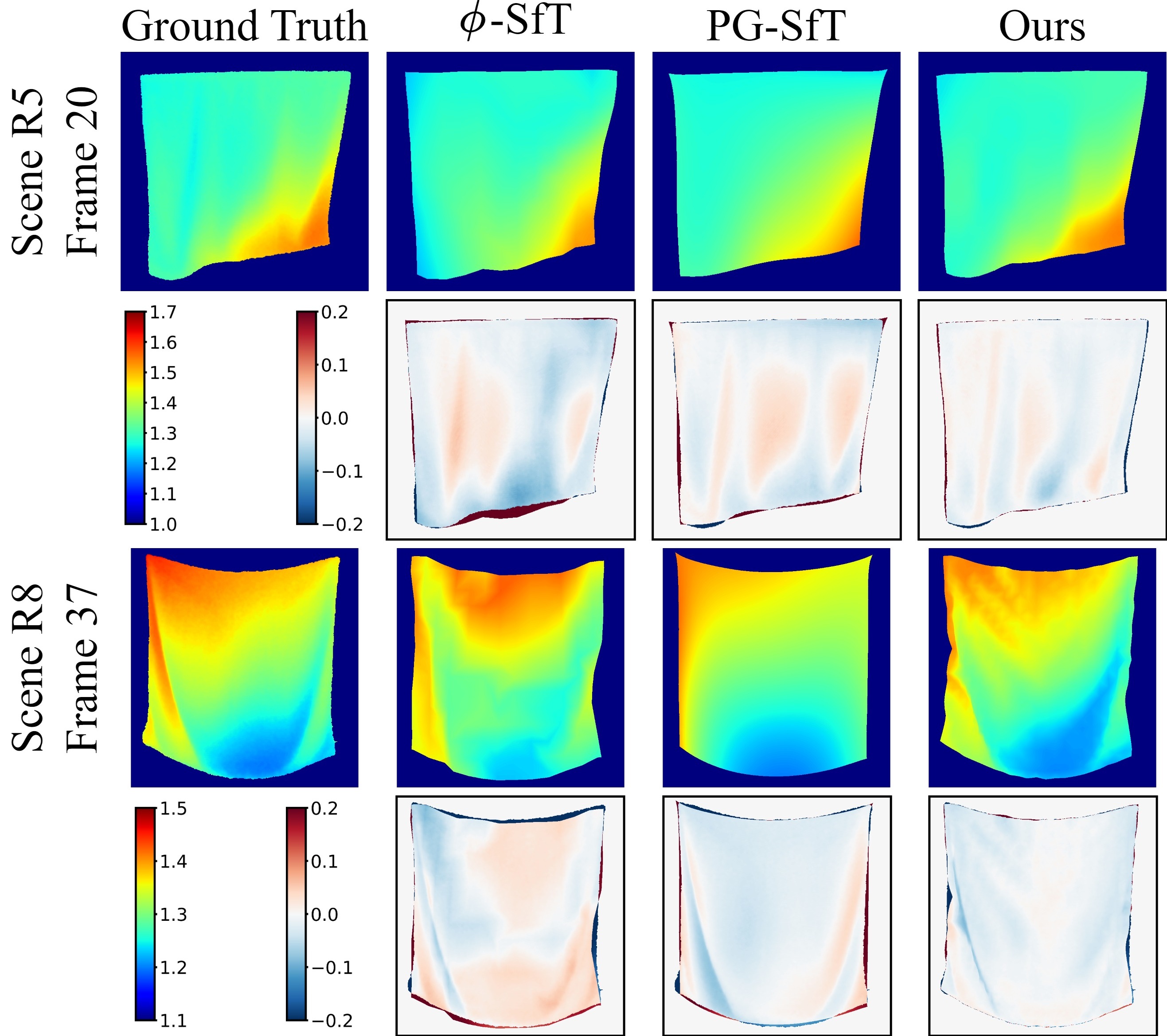}
   \caption{Qualitative comparison between depth images for two example frames of different scenes.}
   \label{figure:depth_comparison}
\end{figure}

\section{Evaluation}
\label{section:evaluation}

\subsection{Qualitative Comparison}

In addition to the qualitative comparison in the main paper, we present mesh renderings and a novel view for the same scenes in \Cref{figure:qualitative_comparison_supplement} and for all scenes in the supplemental video.
The novel view camera is always rotated by $45^\circ$ to the left for all scenes and translated to get a good view on the fabric (with individual translations per scene).

We observe that PG-SfT~\cite{stotko2024physics} produces overly smooth geometries that do not capture high-frequency details (\eg in scenes R2, R3, and R4) and distorted textures due to erroneous $uv$-maps.
$\phi$-SfT~\cite{Phi-SfT} seems to follow the motion well with minor errors.
Especially in scene R2 the renderings of $\phi$-SfT and our method look similar, however, the novel view and the mesh view reveal significant differences on how the large fold looks like.
The fold reconstructed by our method not only looks more realistic, \Cref{section:quantitative_evaluation} shows that the 3D error for our method is much lower in the this scenes compared to the other two approaches indicating the correct folding behavior.
In scene R5 $\phi$-SfT also creates an unrealistic kink close to the top right corner of the mesh which is visible in the novel view in \Cref{figure:qualitative_comparison_supplement} while our method remains smooth as expected for a hanging cloth.
Moreover, the supplemental video displays that our method produces visibly the best accordance with the ground truth video almost all the time.

\Cref{figure:depth_comparison} depicts depth values and depth differences with respect to the ground truth depth data for two scenes.
The depth values show nicely that PG-SfT does not include any fine wrinkles in the reconstructed mesh despite having the highest mesh resolution.
$\phi$-SfT and our method both include such details but our results agree better with the ground truth values than $\phi$-SfT's results.
The supplemental video also points out that our approach yields the best depth values almost all the time.

\subsection{Quantitative Comparison}
\label{section:quantitative_evaluation}

\begin{table*}[h]
    \centering
    \setlength{\tabcolsep}{5.8pt}
    \begin{tabular}{l|ccccccccc|c}
        \toprule
        Method & R1 & R2 & R3 & R4 & R5 & R6 & R7 & R8 & R9 & Mean \\
        \midrule
        $\phi$-SfT \cite{Phi-SfT} & $2.50$ & $1.48$ & $1.29$ & $1.99$ & $2.27$ & $1.71$ & $1.66$ & $1.13$ & $1.01$ & $1.67$ \\
        PG-SfT \cite{stotko2024physics} & $1.60$ & $1.13$ & $2.03$ & $2.08$ & $2.00$ & $2.37$ & $1.54$ & $2.00$ & $1.70$ & $1.83$ \\
        Ours & $\mathbf{0.54}$ & $\mathbf{0.71}$ & $\mathbf{0.76}$ & $\mathbf{1.64}$ & $\mathbf{1.31}$ & $\mathbf{0.81}$ & $\mathbf{1.28}$ & $\mathbf{0.63}$ & $\mathbf{0.51}$ & $\mathbf{0.91}$ \\
        \bottomrule
    \end{tabular}
    \caption{Average depth difference $\delta$ for all methods $[\times 10^{-2} \, \mathrm{m}]$.}
    \label{table:quantitative_comparison_depth_masked}
    \vspace{-4pt}
\end{table*}

\begin{table*}[h]
    \centering
    \setlength{\tabcolsep}{5.8pt}
    \begin{tabular}{l|ccccccccc|c}
        \toprule
        Method & R1 & R2 & R3 & R4 & R5 & R6 & R7 & R8 & R9 & Mean \\
        \midrule
        $\phi$-SfT \cite{Phi-SfT} & $2.21$ & $2.63$ & $2.34$ & $3.56$ & $4.31$ & $3.22$ & $3.04$ & $2.14$ & $1.95$ & $2.82$ \\
        PG-SfT \cite{stotko2024physics} & $2.97$ & $2.14$ & $3.69$ & $4.02$ & $3.64$ & $4.38$ & $3.01$ & $3.79$ & $3.24$ & $3.43$ \\
        Ours & $\mathbf{1.08}$ & $\mathbf{1.24}$ & $\mathbf{1.44}$ & $\mathbf{2.92}$ & $\mathbf{2.57}$ & $\mathbf{1.57}$ & $\mathbf{2.38}$ & $\mathbf{1.22}$ & $\mathbf{1.04}$ & $\mathbf{1.72}$ \\
        \bottomrule
    \end{tabular}
    \caption{Quantitative comparison using the $L_1$ Chamfer distance $[\times 10^{-2} \, \mathrm{m}]$.}
    \label{table:quantitative_chamfer_linear}
    \vspace{-5pt}
\end{table*}

We extend our evaluation by comparing several different metrics for each method.
First, we evaluate the depth images quantitatively by averaging the mean depth errors of all images per scene (see \Cref{section:depth_formula}).
We only use the depth values where ground truth and reconstruction overlap (\ie there are valid depth values to compare).
This excludes small regions in the images in which only one geometry, either the ground truth or the reconstructed one, is present.
We summerize the results in \Cref{table:quantitative_comparison_depth_masked} and observe that our approach beats $\phi$-SfT~\cite{Phi-SfT} and PG-SfT~\cite{stotko2024physics} in all scenes by up to $4.63\times$ and $2.96\times$ respectively.
Both baseline methods have similar performance with mean depth errors of $1.67 \, \mathrm{cm}$ and $1.83 \, \mathrm{cm}$ respectively.
We are able to reduce the average errors by a factor of $1.84$ compared to $\phi$-SfT and $2.01$ compared to PG-SfT.

In addition to the $L_2$ Chamfer distance $\mathrm{CD}_2$, we also report the $L_1$ Chamfer distance $\mathrm{CD}_1$ in \Cref{table:quantitative_chamfer_linear} (\Cref{section:chamfer_formula}).
We use the same point cloud as for the $L_2$ Chamfer distance, that is created by sampling the same number of points as the ground truth point cloud randomly distributed on the reconstructed mesh.
The linear weight for each pair of points reduces the contribution of outliers and focuses on the mean distance between the point clouds instead.
We observe that our approach yields the best result in all scenes again.
On average the $L_1$ Chamfer distance improves by a factor of $1.64$ compared to $\phi$-SfT and by a factor of $1.99$ compared to PG-SfT.

Lastly, we compare the $L_1$ and $L_2$ p2s distance for all methods (see \Cref{section:p2s_formula} for the definitions).
This metric does not include any sampling but only considers the closest point on a triangle instead of including the whole interior.
In 8 out of 9 scenes our method achieves the best results often beating the other approaches by a factor of more than $2.5$ for the $L_1$ distance and a factor of more than $4$ for the $L_2$ distance.
Only in scene R5 PG-SfT performs best.
This may come from the uniform motion in this scene which does not expose the weaknesses of their method.
However, on average our reconstruction outperform $\phi$-SfT~\cite{Phi-SfT} and PG-SfT~\cite{stotko2024physics} by factors of $3.29$ and $2.53$ for the $L_1$ p2s distance and by factors of $3.94$ and $3.43$ for the $L_2$ p2s distance.

\subsection{Fixed Physical Parameters}

We perform a second ablation study that concentrates on the reconstruction quality and stability when the physical material parameters $Y, B$, and $S$ are not optimized but constant at their initial values.
On one hand, these stiffness parameters can not decrease such that the external forces are less likely to become strong enough to crumple the cloth.
On the other hand, the optimization loses its ability to adapt to different fabrics automatically.
We evaluate the $L_2$ Chamfer distances in this scenario with and without our regularization terms and collect the results in \Cref{table:fixed_physical_parameters}.
We observe that fixing the physical parameters mainly prevents complete failures without regularizations.
Similar to the paper's ablation study, some scenes improve in quality due to the variety in deformations but the full model still reaches the best mean error.

\subsection{Real-world Appearance Estimation}

\begin{figure*}
    \begin{subfigure}[t]{0.49\textwidth}
        \centering
        \includegraphics[width=0.99\linewidth]{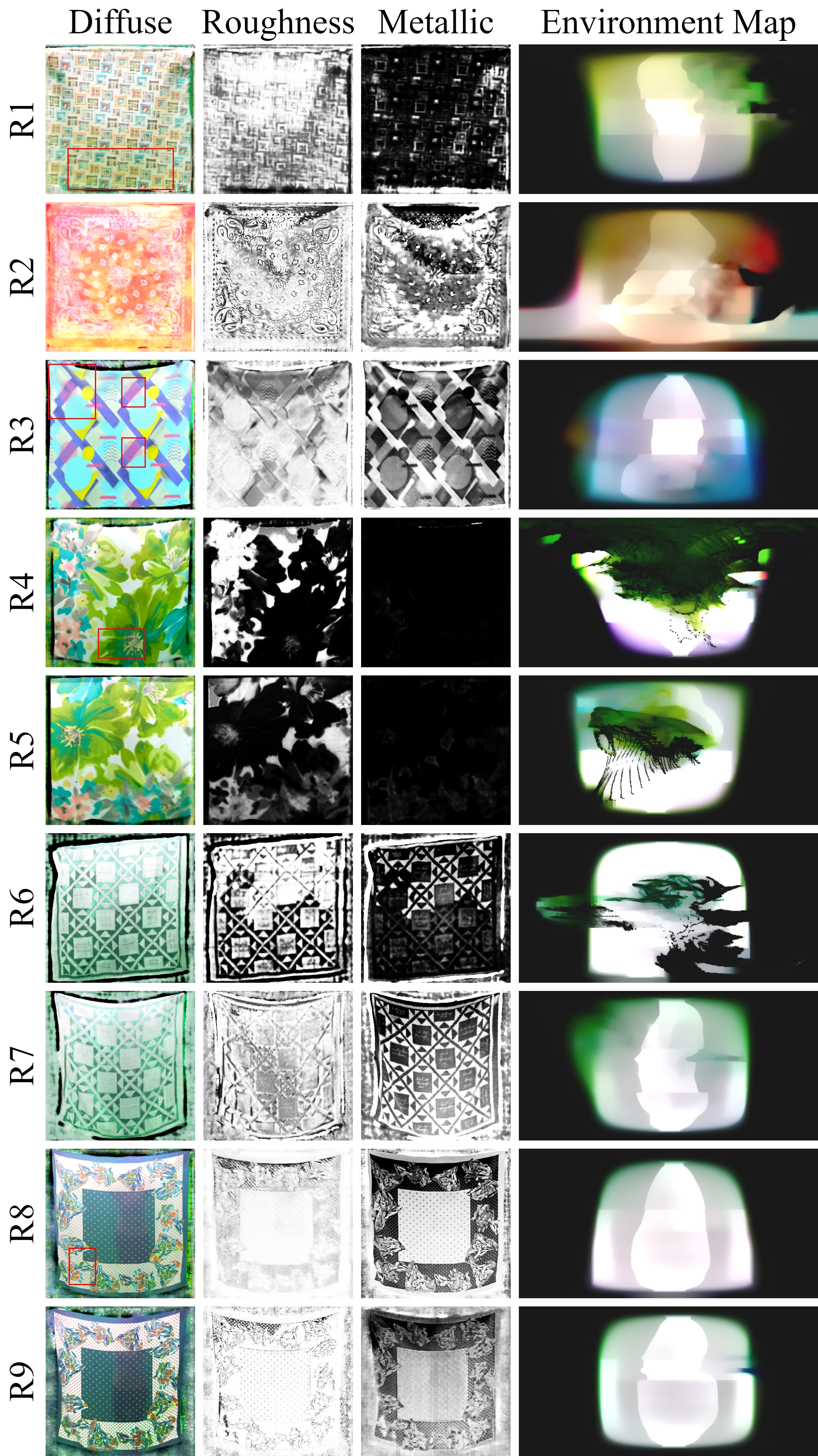}
        \caption{Estimation using the geometry of our 3D reconstruction.}
        \label{figure:material_real}
    \end{subfigure}
    \,
    \begin{subfigure}[t]{0.49\textwidth}
        \centering
        \includegraphics[width=0.99\linewidth]{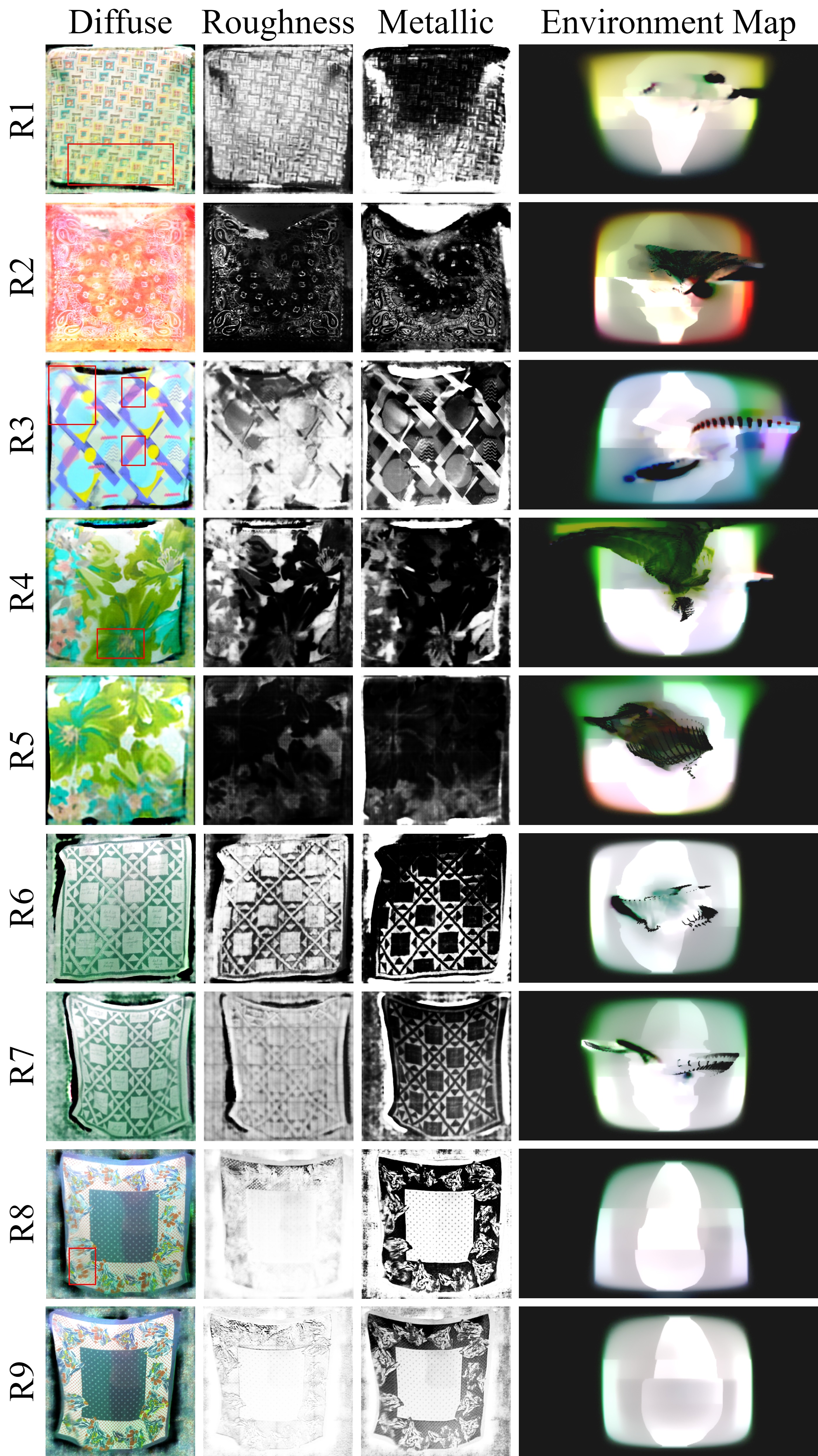}
        \caption{Estimation using the geometry of PG-SfT \cite{stotko2024physics}.}
        \label{figure:material_real_PG_SfT}
    \end{subfigure}
    \caption{Estimated appearance parameters of all real scenes. Major differences are highlighted with a red box. Please zoom in for details.}
\end{figure*}

We complete the evaluation of the optical material estimation by depicting our results for all real-world scenes in \Cref{figure:material_real}.
As a comparison, \Cref{figure:material_real_PG_SfT} shows the results when using the imprecisely reconstructed geometry of PG-SfT \cite{stotko2024physics}.
We emphasize notable differences between the estimates with red rectangles within the diffuse texture in both figures.
Similar characteristics are visible in the roughness and metallic textures.
As described in the main paper, the enhanced geometry reconstruction leads to visibly sharper textures in most of the scenes.
Especially fine structures like the red striped ovals in scene R3 and the flower stamens in scenes R4 and R5 can not be resolved using the geometry of PG-SfT.
The largest difference is visible in the optical material for scene R2.
PG-SfT is not able to reconstruct the deformation at all, causing the appearance optimization to fake the appearance by making the fabric highly reflective.
The environment maps do not show any significant differences because the dominantly diffuse fabrics only make it possible to recover low-frequency features which are very insensitive to the geometry.

\begin{table}[h]
    \centering
    \begin{tabular}{cccccc}
        \toprule
        Scene & SR1 & SR2 & SR3 & SR4 & SR5 \\
        \midrule
        $\mathrm{CD}_2$ & $1.34$ & $1.56$ & $13.7$ & $0.83$ & $0.99$ \\
        \bottomrule
    \end{tabular}
    \caption{3D Reconstruction quality of the synthesized scenes $[\times 10^{-4} \, \mathrm{m}^2]$.}
    \label{table:chamfer_synthetic}
\end{table}

\begin{table*}[h]
    \centering
    \setlength{\tabcolsep}{5.8pt}
    \begin{tabular}{l|ccccccccc|c}
        \toprule
        Method & R1 & R2 & R3 & R4 & R5 & R6 & R7 & R8 & R9 & Mean \\
        \midrule
        $\phi$-SfT \cite{Phi-SfT} & $7.73$ & $3.96$ & $4.45$ & $12.44$ & $21.87$ & $14.08$ & $11.85$ & $\phantom{0}8.25$ & $7.20$ & $10.20$ \\
        PG-SfT \cite{stotko2024physics} & $2.46$ & $5.70$ & $4.94$ & $\phantom{0}7.61$ & $\phantom{0}\mathbf{4.52}$ & $20.12$ & $\phantom{0}4.05$ & $15.26$ & $5.84$ & $\phantom{0}7.83$ \\
        Ours & $\mathbf{0.70}$ & $\mathbf{1.42}$ & $\mathbf{1.58}$ & $\phantom{0}\mathbf{6.80}$ & $\phantom{0}6.94$ & $\phantom{0}\mathbf{2.87}$ & $\phantom{0}\mathbf{3.90}$ & $\phantom{0}\mathbf{1.97}$ & $\mathbf{1.76}$ & $\phantom{0}\mathbf{3.10}$ \\
        \bottomrule
    \end{tabular}
    \caption{$L_1$ point-to-surface distance $\mathrm{p2s}_1$ between the ground truth point cloud and the reconstructed mesh $[\times 10^{-3} \, \mathrm{m}]$.}
    \label{table:quantitative_comparison_p2s}
\end{table*}

\begin{table*}[h]
    \centering
    \setlength{\tabcolsep}{5.8pt}
    \begin{tabular}{l|ccccccccc|c}
        \toprule
        Method & R1 & R2 & R3 & R4 & R5 & R6 & R7 & R8 & R9 & Mean \\
        \midrule
        $\phi$-SfT \cite{Phi-SfT} & $22.65$ & $15.21$ & $\phantom{0}9.71$ & $27.38$ & $62.88$ & $30.26$ & $23.78$ & $10.81$ & $\phantom{0}8.85$ & $23.50$ \\
        PG-SfT \cite{stotko2024physics} & $\phantom{0}3.86$ & $\phantom{0}8.24$ & $11.21$ & $17.51$ & $\mathbf{20.59}$ & $73.12$ & $\phantom{0}6.82$ & $30.55$ & $12.38$ & $20.48$ \\
        Ours & $\phantom{0}\mathbf{0.34}$ & $\phantom{0}\mathbf{1.14}$ & $\phantom{0}\mathbf{2.49}$ & $\mathbf{14.32}$ & $25.26$ & $\phantom{0}\mathbf{2.55}$ & $\phantom{0}\mathbf{4.42}$ & $\phantom{0}\mathbf{1.27}$ & $\phantom{0}\mathbf{1.93}$ & $\phantom{0}\mathbf{5.97}$ \\
        \bottomrule
    \end{tabular}
    \caption{$L_2$ point-to-surface distance $\mathrm{p2s}_2$ between the ground truth point cloud and the reconstructed mesh $[\times 10^{-5} \, \mathrm{m}^2]$.}
    \label{table:quantitative_comparison_p2s_squared}
\end{table*}

\begin{table*}[h]
    \centering
    \setlength{\tabcolsep}{5.8pt}
    \begin{tabular}{l|ccccccccc|c}
        \toprule
        Experiment & R1 & R2 & R3 & R4 & R5 & R6 & R7 & R8 & R9 & Mean \\
            \midrule
            w/o phys. param. & $1.64$ & $4.18$ & $2.31$ & $15.28$ & $\phantom{0}7.32$ & $5.24$ & $\mathbf{3.15}$ & $2.12$ & $1.33$ & $4.73$ \\
            w/o phys. param. \& $\mathcal{R}_F$ & $5.12$ & $5.47$ & $2.29$ & $14.56$ & $\phantom{0}\mathbf{6.83}$ & $5.45$ & $3.29$ & $2.89$ & $1.18$ & $5.23$ \\
            w/o phys. param. \& $\mathcal{R}_E$ & $4.77$ & $1.56$ & $4.02$ & $\phantom{0}9.79$ & $25.65$ & $3.53$ & $4.73$ & $1.72$ & $1.54$ & $6.37$ \\
            w/o phys. param. \& $\mathcal{R}_F$ \& $\mathcal{R}_E$ & $4.52$ & $4.44$ & $\mathbf{2.30}$ & $15.70$ & $\phantom{0}7.42$ & $5.74$ & $3.24$ & $1.58$ & $1.42$ & $5.15$ \\
            Full model & $\mathbf{1.01}$ & $\mathbf{1.55}$ & $2.70$ & $\phantom{0}\mathbf{8.29}$ & $\phantom{0}7.12$ & $\mathbf{2.18}$ & $4.84$ & $\mathbf{1.23}$ & $\textbf{1.04}$ & $\mathbf{3.33}$ \\
        \bottomrule
    \end{tabular}
    \caption{$L_2$ Chamfer distances without optimization of physical material parameters compared to the full model $[\times 10^{-4} \, \mathrm{m}^2]$.}
    \label{table:fixed_physical_parameters}
\end{table*}

\subsection{Synthetic Appearance Estimation}

\begin{figure*}
   \centering
   \includegraphics[width=0.95\linewidth]{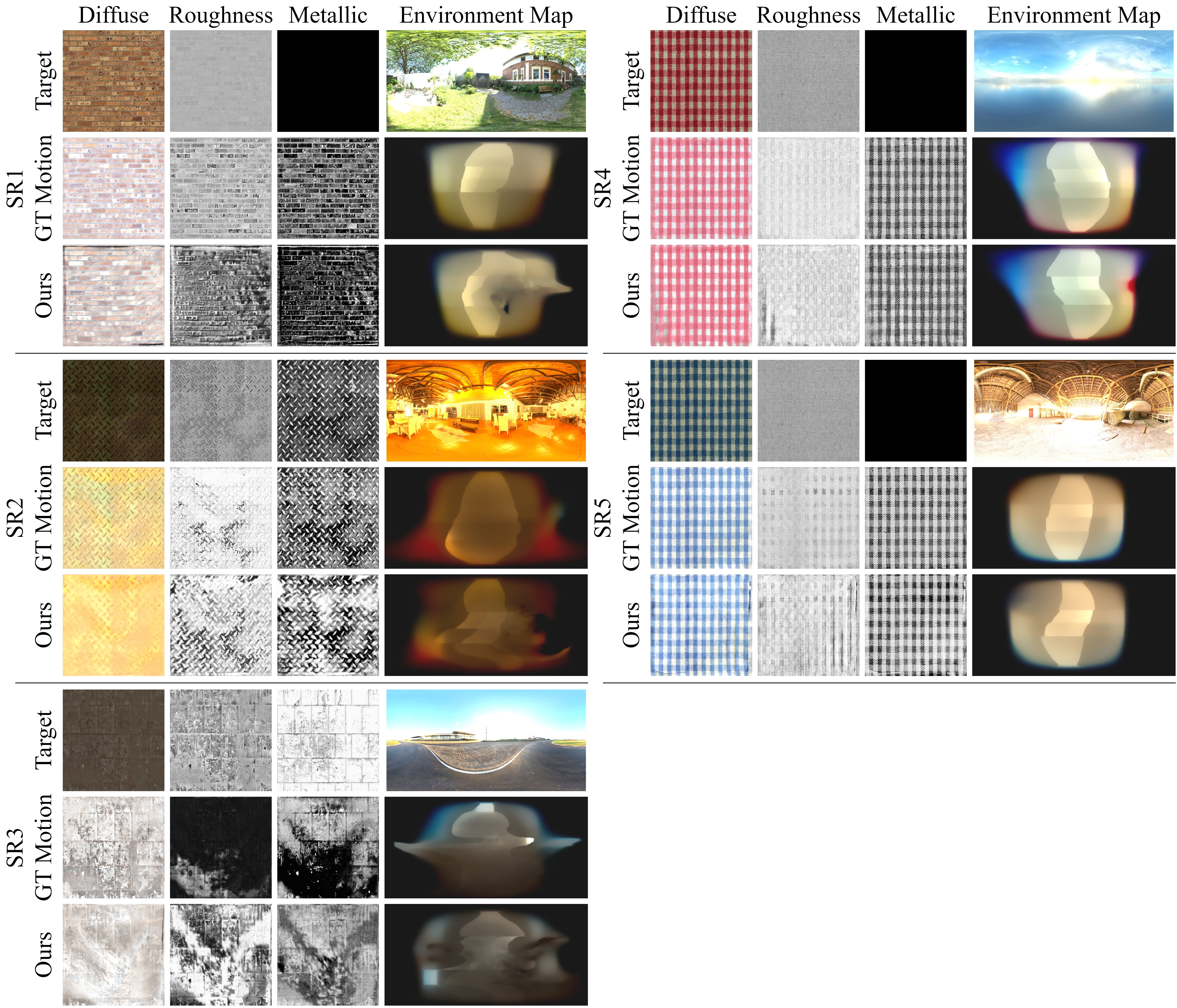}
   \caption{Appearance parameters of all synthesized scenes. The first row depicts the input material textures or the environment map. The second and third row show the estimated appearance maps when using the ground truth geometry or our reconstructed geometry, respectively.}
   \label{figure:material_synthetic}
\end{figure*}

We report the results for the appearance estimation of the synthesized scenes in \Cref{figure:material_synthetic}.
For all five scenes, we depict the target textures, the estimated textures employing the ground truth motion, and the estimated textures using the reconstructed geometry of our method.
The precision of all corresponding geometry reconstructions is given in \Cref{table:chamfer_synthetic}.
All estimates suffer from the ambiguity that different distributions of color and brightness between diffuse texture and environment map may create the same appearance when observing only limited data, such as a single monocular view.
Hence, we mainly compare our results with the ideal case of an estimate based on perfect geometry.
The scenes SR2 and SR3 pose a difficult problem to the geometry reconstruction due to their metallic material causing some specular highlights within the video sequence.
As a consequence, the geometry does not fit the target well enough for scene SR3 (see \Cref{table:chamfer_synthetic}), resulting in blurred textures at the bottom.
Nonetheless, despite little color variations, lots of fine details in the optimized optical material are resolved for scene SR2.
The three non-metallic materials (SR1, SR4, and SR5) show very similar results to the estimates that use the ground truth geometry.
Especially the two fabric materials in scenes R4 and R5 show sharp textures with very little differences.

\end{document}